\documentclass{article}

\usepackage{microtype}
\usepackage{graphicx}
\usepackage{subfigure}
\usepackage{booktabs} 

\usepackage{hyperref}

\usepackage[accepted]{mlsys2025}
\usepackage{macros}
\usepackage[english]{babel}
\usepackage{amsthm}
\usepackage{graphicx, array, blindtext}
\usepackage{dblfloatfix}
\theoremstyle{definition}
\newtheorem{definition}{Definition}

\usepackage{subcaption}
\usepackage{newfloat}
\usepackage{adjustbox}
\usepackage{listings}
\usepackage{xcolor}
\definecolor{mydarkblue}{rgb}{0,0.08,0.45}
\usepackage{cleveref}
\usepackage{tikz}
\usepackage{mathtools} 
\usepackage{amsfonts} %
\usepackage{pgfplots}
\pgfplotsset{compat=1.18}
\usetikzlibrary[pgfplots.groupplots]
\usepackage{wrapfig}
\usepackage{booktabs}

\usepackage[symbol]{footmisc}

\usetikzlibrary{arrows, automata, positioning}
\tikzset{auto, >=stealth}
\tikzset{every edge/.append style={shorten >= 1pt}}
\tikzset{
    main node/.style={circle,draw,minimum size=1cm,inner sep=0pt},
}

\DeclareCaptionStyle{ruled}{labelfont=normalfont,labelsep=colon,strut=off} 
\floatstyle{ruled}
\newfloat{listing}{tb}{lst}{}
\floatname{listing}{Listing}
\mlsystitlerunning{Know Where You’re Uncertain When Planning with Multimodal Foundation Models: A Formal Framework}

\begin{document}

\twocolumn[
\mlsystitle{Know Where You’re Uncertain When Planning with \\ Multimodal Foundation Models: A Formal Framework}



\mlsyssetsymbol{equal}{*}

\begin{mlsysauthorlist}
\mlsysauthor{Neel~P.~Bhatt}{equal,to}
\mlsysauthor{Yunhao~Yang}{equal,to}
\mlsysauthor{Rohan~Siva}{to}
\mlsysauthor{Daniel~Milan}{to}
\mlsysauthor{Ufuk~Topcu}{to}
\mlsysauthor{Zhangyang~Wang}{to}
\end{mlsysauthorlist}

\mlsysaffiliation{to}{The University of Texas at Austin, United States}

\mlsyscorrespondingauthor{Neel~P.~Bhatt}{npbhatt@utexas.edu}
\mlsyscorrespondingauthor{Yunhao~Yang}{yunhaoyang234@utexas.edu}

\mlsyskeywords{Uncertainty Estimation, Foundation Models, Multimodal, Planning, Conformal Prediction, Formal Methods}

\vskip 0.3in

\begin{abstract}

Multimodal foundation models offer a promising framework for robotic perception and planning by processing sensory inputs to generate actionable plans. However, addressing uncertainty in both perception (sensory interpretation) and decision-making (plan generation) remains a critical challenge for ensuring task reliability. We present a comprehensive framework to disentangle, quantify, and mitigate these two forms of uncertainty. We first introduce a framework for uncertainty \textit{disentanglement}, isolating \textit{perception uncertainty} arising from limitations in visual understanding and \textit{decision uncertainty} relating to the robustness of generated plans.

To quantify each type of uncertainty, we propose methods tailored to the unique properties of perception and decision-making: we use conformal prediction to calibrate perception uncertainty and introduce Formal-Methods-Driven Prediction (FMDP) to quantify decision uncertainty, leveraging formal verification techniques for theoretical guarantees. Building on this quantification, we implement two targeted \textit{intervention} mechanisms: an active sensing process that dynamically re-observes high-uncertainty scenes to enhance visual input quality and an automated refinement procedure that fine-tunes the model on high-certainty data, improving its capability to meet task specifications. Empirical validation in real-world and simulated robotic tasks demonstrates that our uncertainty disentanglement framework reduces variability by up to 40\% and enhances task success rates by 5\% compared to baselines. These improvements are attributed to the combined effect of both interventions and highlight the importance of uncertainty disentanglement which facilitates targeted interventions that enhance the robustness and reliability of autonomous systems. Webpage, videos, demo, and code: \href{https://uncertainty-in-planning.github.io/}{https://uncertainty-in-planning.github.io}.

\end{abstract}

]



\printAffiliationsAndNotice{\mlsysEqualContribution} 

\section{Introduction}\label{sec:intro}

Accurate perception and the generation of actionable plans are essential for effective robotic perception and planning \citep{llm-planner, lm-nav, yang2023multimodal, AutomatonBasedRO, saycan, bhatt2023mpc, wang2024planning}. Recent advancements in multimodal foundation models have equipped robots with the ability to process visual data and generate corresponding textual plans \citep{wu2023embodied, gu2023conceptgraphs, mu2023embodiedgpt}. While multimodal models provide integrated perception and planning capabilities, these two components — perception and decision-making — each contribute distinct forms of uncertainty that affect overall performance. Without disentangling these sources, any observed failures or inconsistencies in model outputs remain difficult to diagnose, as it is unclear whether they stem from inaccuracies in visual perception or limitations in planning. 

Disentangling uncertainty allows for targeted improvements: enhancing perception uncertainty focuses on refining sensory interpretation and visual recognition, whereas addressing decision uncertainty improves the robustness and alignment of generated plans to task specifications. By separately identifying and mitigating these uncertainties, we may enable targeted interventions that enhance model reliability and adaptability in dynamic, real-world environments. 

This paper aims to bridge this largely unaddressed gap \citep{ye2023uncertainty, vazhentsev2022uncertainty, knowno2023} by introducing a novel framework to distinguish and quantify the uncertainty associated with each constituting module, thereby enabling targeted model refinement and improved execution reliability. Specifically:
\begin{itemize}\vspace{-0.5em}
    \item \textit{\textbf{Perception uncertainty}} arises from the model’s limitations in visually understanding and reliably interpreting sensory data, such as recognizing and localizing objects. For instance, in a cluttered room, high perception uncertainty can manifest in the robot's inability to accurately detect and localize obstacles in its path.
    \item \textit{\textbf{Decision uncertainty}} pertains to the model’s planning capabilities, specifically its capacity to generate a sequence of actions that aligns with the task requirements. Using the same navigation example, decision uncertainty would reflect the model’s ability to formulate a plan that avoids all obstacles and successfully reaches the target destination. Importantly, decision uncertainty can persist even when visual perception is perfect, highlighting the need for separate evaluation and mitigation of each uncertainty source.\vspace{-0.5em}
\end{itemize}

\textbf{Contributions.} This paper advances multimodal foundation models in robotic planning through three key contributions:\vspace{-5pt}

\begin{itemize}
    \item \textbf{Uncertainty Disentanglement Framework}: We introduce a novel framework to disentangle \textit{perception uncertainty} and \textit{decision uncertainty} within multimodal foundation models, addressing the unique challenges each type of uncertainty presents in robotic planning. This disentanglement allows us to identify and treat these uncertainties separately, enabling targeted improvements in model robustness.\vspace{-0.2em}
    \item \textbf{Novel Quantification of Each Uncertainty}: To estimate these uncertainties, we propose novel methods tailored to each type. For perception uncertainty, we leverage \textit{conformal prediction} to calibrate the model’s visual confidence, providing a probabilistic measure of the model’s accuracy in object recognition. For decision uncertainty, we introduce \textit{Formal-Methods-Driven Prediction} (FMDP), which uses formal verification techniques alongside conformal prediction to assess the likelihood that generated plans will satisfy task-specific requirements, offering a robust means to validate planning outputs.\vspace{-0.2em}
    \item \textbf{Targeted Interventions via Active Sensing and Automated Refinement}: Building on this uncertainty quantification, we implement a two-part improvement strategy. First, our active sensing mechanism dynamically initiates re-observation of high-uncertainty scenes, enhancing visual input quality and reducing the propagation of perceptual errors. Second, we introduce an uncertainty-aware fine-tuning procedure that refines the model on low-uncertainty samples, improving its consistency in meeting task specifications while eliminating the need for human annotations.\vspace{-5pt}
\end{itemize}

We empirically validate the proposed frameworks through experiments in ground robot navigation, both in real-world environments and using Carla simulations \cite{Dosovitskiy17}. Unlike conventional refinement frameworks, which lack the precision to identify and address specific uncertainty sources, our disentangled approach achieves a significant reduction in variability of model outputs — up to \textbf{40\%} — and enhances the probability of satisfying task specifications by up to \textbf{5\%}. These enhancements are a direct result of the synergistic combination of our targeted interventions and highlight the importance of distinguishing between sources of uncertainty in autonomous systems, ultimately improving their reliability and adaptability.

\vspace{-0.1em}
\subsection{Related Work} 
\vspace{-0.2em}
Multimodal foundation models, such as GPT-4 \citep{gpt-4} and LLaVA \citep{liu2023llava, liu2023improvedllava}, have advanced robotic systems by jointly processing text and images to generate actionable plans. Existing uncertainty quantification methods \citep{ye2023uncertainty, vazhentsev2022uncertainty, shen2021real, knowno2023, hori2023interactively} treat these models as monolithic, providing an aggregate measure of uncertainty without distinguishing between perception and decision components. This results in a “black box” evaluation, where any assessment of task failure or risk lacks insight into whether uncertainty originates from perception limitations or decision-making flaws.

Such aggregate uncertainty scores complicate model refinement since they obscure the root cause of performance issues, hindering targeted improvements \cite{rudner2024fine, yu2022actune, jang2024online}. For instance, while log-likelihood-based uncertainty estimation methods \citep{srivastava2022beyond, hendrycks2020measuring} use conformal prediction techniques \citep{conformal-prediction, angelopoulos2023conformal, conformalML} to calibrate confidence in plan correctness, they often rely on human-labeled ground truth for accuracy. This dependence can limit scalability and adaptability in dynamic environments.

Our framework addresses these limitations by disentangling uncertainty into perception and decision components, enabling targeted interventions to reduce each type of uncertainty independently. The proposed FMDP estimates decision uncertainty without requiring extensive human labeling. This disentanglement and targeted quantification significantly enhance model robustness and scalability through targeted uncertainty mitigation.

\section{Preliminary Background}

\paragraph{Formal Methods in Decision-Making}
Formal methods provide tools for modeling, analyzing, and verifying decision-making problems represented mathematically. By representing model outputs in a symbolic representation such as a Kripke structure, $\mathcal{A}$, these methods provide formal verification of the model's output against task specifications \cite{browne1988characterizing}.

We use temporal logic to express the task specifications. A temporal logic specification, $\phi$, constrains the temporal ordering and logical relationship between sequences of events and actions. Subsequently, we perform model checking, a process to determine whether the Kripke structure satisfies the specification, denoted as $\mathcal{A} \models \phi$, which provides a formal guarantee associated with this verification.

\paragraph{Conformal Prediction}
Through the theory of conformal prediction, we can estimate the uncertainty of model predictions and obtain statistical guarantees on the correctness of these predictions \citep{conformal-prediction}.

Formally, consider a space of (data, truth label) pairs $\mathcal{X} \times \mathcal{Y}$. We have a \emph{calibration set} $S_C = \{(x_1, y_1),..., (x_n, y_n)\}$ with $n$ elements from the space $\mathcal{X} \times \mathcal{Y}$. 
These pairs are independent and identically distributed (i.i.d.).
The goal is to find a \emph{prediction band} $\hat{C}: \mathcal{X} \rightarrow \{\text{subsets of } \mathcal{Y}\}$. For example, $\hat{C}(x_i) = \{y_1, ...,y_i\} \subset \mathcal{Y}$.

Given a new test pair $(x_{n+1}, y_{n+1})$ sampled from $\mathcal{X} \times \mathcal{Y}$, we first make the following assumption:

\textbf{Assumption 1.} $(x_{n+1}, y_{n+1})$ is sampled from a distribution that is identically distributed with the distribution from which we sampled $S_C= \{(x_1, y_1),..., (x_n, y_n)\}$.

Let $\epsilon \in [0, 1]$ be a score reflecting uncertainty. If assumption 1 holds, we expect the prediction band to meet the property
\begin{equation}
    \label{eq: cp}
    \mathbb{P}\left[y_{n+1} \in \hat{C}(x_{n+1}) \right] \ge 1 - \epsilon.
\end{equation}
The objective is to find $\hat{C}$ such that the probability of the ground truth label $y_{n+1}$ belonging to the set $\hat{C}(x_{n+1})$ is bounded by $1-\epsilon$. For instance, if $\hat{C}(x_{n+1})$ consists of two elements, then the probability of at least one of the elements matching the ground truth label is $1 - \epsilon$.

\paragraph{Table of Notations} For convenience, we present all notations used in the rest of the paper in Table \ref{tab: notations}.
\begin{table}[t]
    \centering
    \begin{tabular}{ccc}
    \toprule
        Notation & Meaning & First Def. \\
    \midrule
        M & foundation model & Sec. \ref{sec: perception-unc} \\
        $\mathcal{T}, \mathcal{I}$ & text, image space &  Sec. \ref{sec: perception-unc} \\
        $I$ or $I_i$ & a single image & Sec. \ref{sec: perception-unc} \\
        $T$ or $T_i$ & generated text instruction & Sec. \ref{sec: perception-unc} \\
        $V$ & vision encoder & Sec. \ref{sec: conf-perc-unc-calibrate} \\
        $H$ & projection head & Sec. \ref{sec: conf-perc-unc-calibrate} \\
        $c$ & softmax confidence score & Sec. \ref{sec: conf-perc-unc-calibrate} \\
        $y$ or $y_i$ & image object label & Sec. \ref{sec: conf-perc-unc-calibrate} \\
        $u_p$ & percep. uncertainty score & Def. \ref{def: perception} \\
        $t_p$ & percep. threshold & Sec. \ref{sec: active-sensing} \\
        $P_T$ or $P_{T_i}$ & textual task description & Sec. \ref{sec: decision-unc} \\
        $Y$ or $Y_i$ & a set of observed objects & Sec. \ref{sec: fmdp} \\
        $u_d$ & decision uncertainty score & Def. \ref{def: decision} \\
        $t_d$ & decision threshold & Sec. \ref{sec: estimate-dec-unc} \\
    \bottomrule
    \end{tabular}
    \caption{A summary of important notations.}
    \label{tab: notations}
\end{table}
\begin{figure*}[t]
    \centering
    \includegraphics[width=0.9\linewidth]{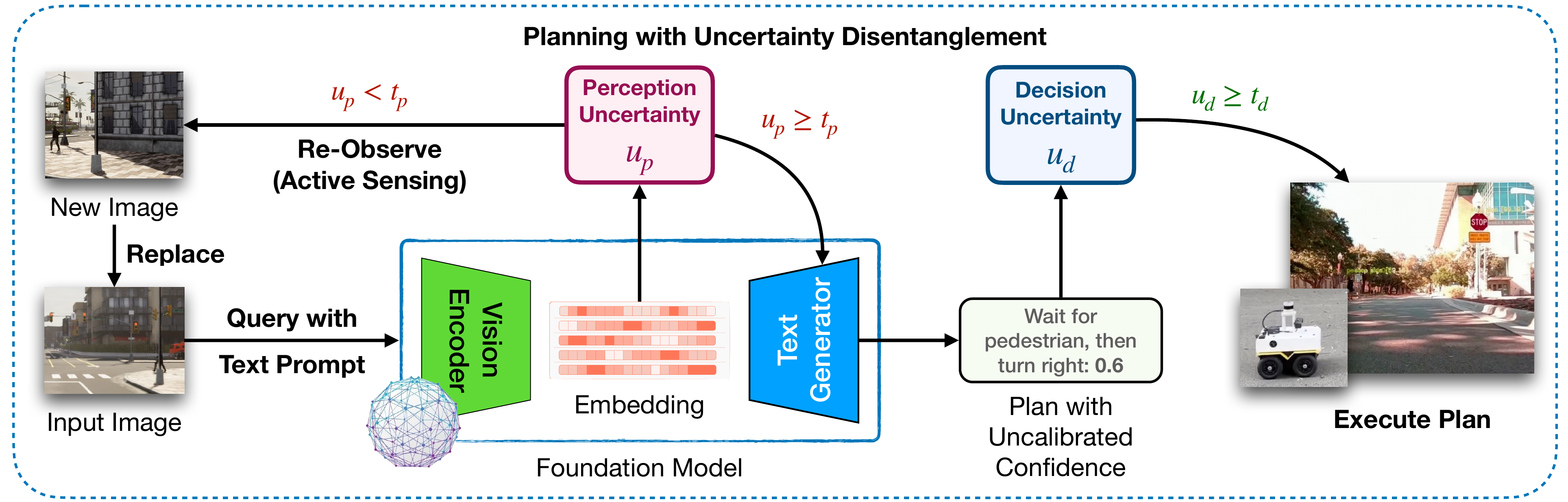}
    \caption{Our planning framework disentangles perception and decision uncertainty, triggering the active sensing intervention. The framework improves the robustness of generated plans by reducing the propagation of perceptual inaccuracies.}
    \label{fig: strategy}
    \vspace{-10pt}
\end{figure*}

\section{Perception Uncertainty}
\label{sec: perception-unc}
A multimodal foundation model,  $M: \mathcal{I} \times \mathcal{T} \rightarrow \mathcal{T} \times [0,1]$, takes an image observation $I$ in the image space $\mathcal{I}$, and a task description $P_T$ in the text space $\mathcal{T}$ as inputs. It then produces a textual instruction $T \in \mathcal{T}$ for the given task and a softmax confidence $c \in [0,1]$.  $M$ consists of a vision encoder $V: \mathcal{I} \rightarrow \mathbb{R}^{d}$ that maps input images from the image space to a d-dimensional embedding space. The vision encoder outputs a d-dimensional vector which is sent as an input to a text generator for producing textual instructions. 
We define \emph{perception uncertainty} in the vision encoder's output as below:
\begin{definition}
    \label{def: perception}
    Let $I$ be an image containing multiple objects, a \textit{perception uncertainty score} $u_p$ for $I$ is a theoretical lower bound on the probability of correctly identifying all objects in the image. A lower score implies higher uncertainty.
\end{definition}

\textbf{Problem 1.} Given a foundation model $M$, a task description $P_T \in \mathcal{T}$ with an image $I \in \mathcal{I}$, estimate the \textit{perception uncertainty score} $u_p$ for $I$.

\subsection{Conformal Confidence-Uncertainty Calibration}
\label{sec: conf-perc-unc-calibrate}
As ad-hoc confidence scores returned by machine learning models, such as softmax, do not accurately reflect prediction accuracies \cite{guo2017calibration}, it is essential to calibrate confidence scores to prediction accuracies for uncertainty estimation. However, we ask for a theoretical lower bound on the accuracy according to Def. \ref{def: perception}, which existing calibration methods cannot provide \cite{guo2017calibration, temperature}. Conformal prediction provides confidence intervals or sets for predictions guaranteeing a specified error rate \cite{conformal-prediction}. We leverage the theory of conformal prediction for confidence-uncertainty calibration.

We first learn a \emph{projection head} $H: \mathbb{R}^d \rightarrow \mathbb{R}^k$ that maps image embeddings to a $k$-dimensional space where vector $v \in \mathbb{R}^k$ represents the softmax confidence scores over $k$ object classes. We denote the confidence score for class $l \in [0, k]$ by $v_l$. The class associated with the highest confidence score is the model's \emph{prediction}.

Consider a calibration set $\{I_i, y_i\}_{i=1}^n$, where $I_i \in \mathcal{I}$ and $y_i \in [0, k]$ is the true class label for the object of interest in $I_i$, and assume the calibration set is i.i.d. with the images we from evaluation tasks.
We connect the vision encoder to the projection head $H \circ V: \mathcal{I} \rightarrow \mathbb{R}^k$ and obtain a set of \emph{nonconformity scores} $S_{nc} = \{1 - H(V(I_i))_{y_i}\}_{i = 1}^n$. Intuitively, a nonconformity score is the sum of the softmax confidences of wrong predictions.

We empirically estimate a probability density function (PDF) $f_{nc}$ from the nonconformity scores. For an image $I_{n+1}$ from the test task (beyond the calibration set), given an error bound $\epsilon$, we can find corresponding confidence $c^*$ such that $\hat{C}(I_{n+1}) = \{l\ : \ H(V(I_{n+1}))_l > 1 - c^* \}$ satisfies $\mathbb{P}\left[y_{n+1} \in \hat{C}(I_{n+1}) \right] \ge 1 - \epsilon$ \citep{conformal-prediction}.

Conversely, given a $c^*$, we can compute the corresponding error bound $\epsilon = 1 - \int_0^{c^*} f_{nc}(x) dx$. Given a confidence vector $v \in \mathbb{R}^k$ of image $I_{n+1}$, let $c^* = 1 - sort(v)_{-2}$, where $sort(v)_{-2}$ is the second-greatest value in $v$. The prediction band becomes
\begin{equation}
    \label{eq: pred-band}
    \begin{split}
        \hat{C}(I_{n+1}) & = \{l\ : \ H(V(I_{n+1}))_l > 1 - c^* \} \\
        & = \{l\ : \ H(V(I_{n+1}))_l > sort(v)_{-2} \} \\
        & = \{ argmax ( H(V(I_{n+1})) ) \} .
    \end{split}
\end{equation}
The prediction band contains a single element, which is the prediction with the maximum confidence. Let $y_{n+1} \in [0, k]$ be the ground truth object label. By the theory of conformal prediction, there exists an $\epsilon \in [0,1]$ such that
\begin{equation}
    \begin{split}
        \mathbb{P} \left[ y_{n+1} \in \hat{C}(I_{n+1}) \right] = & \\ \mathbb{P} \left[ y_{n+1} = argmax ( H(V(I_{n+1})) \right] & \ge 1 - \epsilon,
    \end{split}
\end{equation}
where $1 - \epsilon$ is the perception uncertainty score per Def. \ref{def: perception}.

We know that $c^*$ is the $1-\epsilon$ quantile of $S_{nc}$. Thus, $1 - \epsilon = \int_0^{c^*} f_{nc}(x) dx$. Then,
\begin{equation}
    \label{eq: c-epsilon}
    \mathbb{P}\left[ y_{n+1} = argmax ( H(V(I_{n+1})) \right] \ge 1- \epsilon = \int_0^{c^*} f_{nc}(x) dx.
\end{equation}
Hence, we can calibrate a confidence vector $v$ into a perception uncertainty score
\begin{equation}
    \label{eq: perception-unc-score}
    u_p = \int_0^{c^*} f_{nc}(x) dx, \text{ where } c^* = 1 - sort(v)_{-2}.
\end{equation}

\section{Decision Uncertainty}
\label{sec: decision-unc}

Existing frameworks depend on human-annotated datasets to assess how well a generated plan aligns with human annotations, limiting their scalability \cite{malinin2017incorporating, jiang2021can}.
These approaches are limited in their ability to adapt to specific task requirements, resulting in less effective uncertainty estimates when applied to diverse or specialized scenarios \cite{knowno2023}. We use formal verification techniques alongside conformal prediction to eliminate the need for human annotations and improve adaptability to various task domains.

Recall that a foundation model $M: \mathcal{I} \times \mathcal{T} \rightarrow \mathcal{T} \times [0,1]$ returns a text-based instruction to complete a task and a confidence score between 0 and 1. For example,
\vspace{4pt}
\begin{lstlisting}[language=completion]
    <prompt><Image> <Task description></prompt>
    <completion><Instruction></completion>
    <prompt>Does the instruction satisfy the rules?
    <Specifications></prompt>
    <completion>Y/N</completion>
\end{lstlisting}
The inputs to $M$ are in blue and the outputs are in red. The confidence refers to the softmax score of token `Y' (Yes).

Given a set $\Phi$ of temporal logic specifications, we expect the text-based instructions generated by $M$ to satisfy all specifications. However, due to the black-box nature of $M$, we cannot provide a guarantee. Thus, we aim to estimate a \emph{decision uncertainty score} for each generated instruction.

\begin{definition}
    \label{def: decision}
    Let $\Phi$ be a set of temporal logic specifications and $T$ be a text-based instruction. A \textit{decision uncertainty score} $u_d$ for $T$ is a numeric value estimating the probability that $T$ satisfies all specifications $\phi \in \Phi$.
\end{definition}

We formulate our problem as:

\textbf{Problem 2.} Given a foundation model $M$, a set $\Phi$ of specifications, a task description $P_T \in \mathcal{T}$, and an image $I \in \mathcal{I}$, estimate the \textit{decision uncertainty score} $u_d$ for instruction $T$, where $T, c = M(P_T, I)$.

\subsection{Formal-Methods-Driven Prediction}
\label{sec: fmdp}

We present \textbf{formal-methods-driven prediction (FMDP)}, which uses formal verification techniques alongside conformal prediction, to solve Problem 2. For each task description $P_T$ and input image $I$, FMDP aims to find a prediction band $\hat C (M, P_T, I) \subseteq \{T \ |\ T, c = M(P_T, I)\}$ such that
\begin{equation}
\label{eq: fmdp}
    \begin{split}
        \mathbb{P} & \left[ T \in \hat C ( M, P_T, I ) : \forall_{\phi \in \Phi} \ \textbf{pEnc}(T, AP, Y) \models \phi \right] 
        \\ & \ge 1 - \epsilon
    \end{split}
\end{equation}
for a given error tolerance $\epsilon \in [0, 1]$. 
$Y$ is the set of objects from $I$, \textbf{pEnc} is an algorithm converting text to automaton, defined in Alg. \ref{alg: text2automata}.
$AP$ is the set of atomic propositions that are used to formulate every $\phi \in \Phi$. Eq. \ref{eq: fmdp} implies that the probability that ``at least one of the instructions obtained using the prediction band $\hat C$ satisfies the constraints defined by $\Phi$'' is at least $1-\epsilon$.

\paragraph{Text-Based Instruction to Automaton}
Problem 2 requires verifying the textual outputs from $M$ against the specifications.
As we cannot directly verify texts against a logical specification, we develop an algorithm that expresses text-based instruction in a formal representation. The algorithm takes a textual instruction $T$, a set $AP$ of atomic propositions, and a set $Y$ of objects observed from them as inputs. It returns a \emph{Kripke structure} \cite{browne1988characterizing} to represent the instruction. We show the details in Alg. \ref{alg: text2automata}. 

\begin{definition}
    A Kripke structure $\mathcal{A} = (\AutStates, Q_0, \AutTransFunc, \AutLabelFunc)$ is a tuple consisting of a set $\AutStates$ of states, a set $Q_0$ of initial states, transitions $\AutTransFunc \subseteq \AutStates \times \AutStates$, and a label function $\AutLabelFunc: \AutStates \rightarrow 2^{AP}$.
\end{definition}

In particular, the algorithm first parses the textual instruction into a set of verb and noun phrases using semantic parsing \cite{semantic-parsing}. Next, we create an initial state $q_0$ whose label is the intersection of the set of atomic propositions and the set of observed objects. Then, we create a set of states, where each state corresponds to a phrase, and connect them linearly. Finally, we create a ``final'' state indicating the completion of the task. We present a step-by-step running example in Appendix \ref{sec: algo-demo}.

Then, we verify this structure against the specification through a model checker \cite{model-checking}. 
The model checker takes a logical specification and a Kripke structure as inputs and returns a binary signal indicating whether this structure satisfies the specification. We consider the textual instruction to satisfy the specification if and only if its corresponding Kripke structure passes the model-checking step.

\begin{algorithm}[t]
  \caption{Natural Language to Kripke Structures}\label{alg: text2automata}
  \begin{algorithmic}
    \STATE\algorithmicinput{ textual instruction $T$, atomic proposition set $AP$, set $Y$ of observed objects}
    \STATE\algorithmicoutput{ $(\AutStates, q_0, \AutTransFunc, \AutLabelFunc)$} 

    \STATE $Ph = \{ Ph_1, Ph_2, ... \}$ = \textbf{parse}($T$)

    \STATE $\AutStates, \AutTransFunc$ = [$q_0$], [] \hfill\COMMENT{Define a set of states and transitions. $q_0$ denotes initial states}
    \STATE $\AutLabelFunc (q_0) = Y \cap AP$ \hfill\COMMENT{The initial state's label is the observed objects from the image}

    \STATE\algorithmicfor{ $Ph_i$ in $Ph$}
        \STATE $\quad \AutStates$.append($q_i$), $\AutTransFunc$.append($(q_{i-1}, q_i)$), 
        \STATE $\quad \AutLabelFunc (q_i) = \{ p \in AP : p \in Ph_i \}$
    \STATE\algorithmicendfor
    \STATE $\AutStates$.append($q_{done}$), $\AutTransFunc$.append($(q_{|Ph|}, q_{done}))$, 
    \STATE $\AutTransFunc$.append($q_{done}, q_{done}))$, $\AutLabelFunc(q_{done}) = \emptyset$
  \end{algorithmic}
\end{algorithm}

\paragraph{Calibration Set and Nonconformity Distribution}
Once we enable the verification of the generated instructions, we can use the verification outcomes to formulate a calibration set for estimating a nonconformity distribution.

Given a set $\{I_j, Y_j\}_{j=1}^n$ of images with observed objects, e.g., the objects for the second image in Fig. \ref{fig: uncertainty-case-study} are $\{$truck, car$\}$, we supply each image $I_j$ with a task description $P_{T_j} \in \mathcal{T}$ to $M$ and obtain an instruction $T_j$ and a confidence score $c_j$: $T_j, c_j = M(P_{T_j}, I_j)$. By repeating this step for all images, we obtain our calibration set $\{(T_j, c_j, Y_j)\}_{j=1}^n$.

We then obtain a set $S_{nc}$ of \emph{nonconformity scores} 
\begin{equation}
    S_{nc} = \left\{
    1 - c_j \ :\ \exists_{\phi \in \Phi} \ \textbf{pEnc}(T_j, AP, Y_j) \models \phi 
    \right\}_{j=1}^n.
\end{equation}
We use $S_{nc}$ to estimate a PDF $f_{nc}$ for the nonconformity distribution.

After we get the set of nonconformity distributions, for a given error bound $\epsilon \in [0, 1]$, we can find a prediction band $\hat C (M, P_T, I) = \{T \ |\ T, c = M(P_T, I) \text{ and } c \ge 1 - c^*\}$ that satisfies Eq. \ref{eq: fmdp}, where $c^*$ is the $1-\epsilon$ quantile of $S_{nc}$.

\begin{figure*}[t]
    \centering
    \includegraphics[width=0.9\linewidth]{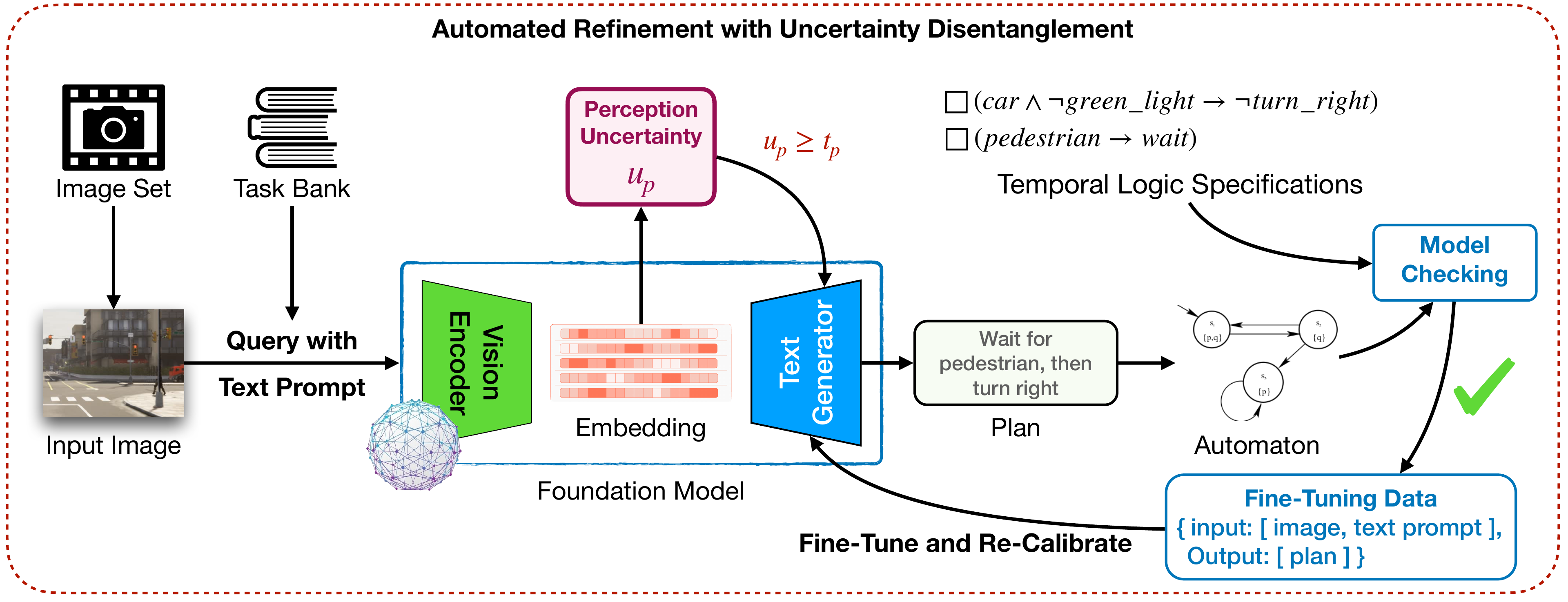}
    \caption{Our automated refinement framework generates high-certainty training data and fine-tunes the foundation model to improve its ability to generate plans that comply with task requirements.}
    \vspace{-10pt}
    \label{fig: fine-tune-pipeline}
\end{figure*}

\subsection{Estimating Decision Uncertainty Score}
\label{sec: estimate-dec-unc}
In this section, we use FMDP to estimate a decision uncertainty score for each output from the foundation model, i.e., solve Problem 2.

Given a new image $I_{n+1}$ and a task description $P_{T_{n+1}}$, the foundation model outputs an instruction $T_{n+1}$ and a confidence $c_{n+1}$. We disregard the instructions whose confidence in satisfying the specifications is below 0.5.
Then, we compute the decision uncertainty score $u_d$ using the nonconformity distribution $f_{nc}$:
\begin{equation}
    \label{eq: decision-unc-score}
    u_d = \int_0^{c_{n+1}} f_{nc}(x) dx.
\end{equation}
\begin{proof}
    Consider a prediction band $\hat C (M, P_T, I) \subseteq \{T \ |\ T, c = M(P_T, I) \text{ and } c \ge 1 - c^*\}$ that satisfies Eq. \ref{eq: fmdp}. For each $\epsilon$, there is a corresponding value of $c^*$.
    Let $c^* = c$, $\hat C (M, P_T, I)$ will contain only one element, which is the instruction $T$. By the theory of conformal prediction, there exists an $\epsilon$ such that
    \begin{center}
        $\mathbb{P} \left[ T : \forall_{\phi \in \Phi} \ \textbf{pEnc}(T, AP, Y) \models \phi \right] \ge 1 - \epsilon$.
    \end{center}
    The left-hand side is our decision uncertainty score $u_d$ according to Def. \ref{def: decision}. Therefore, if we find the value of $\epsilon$, we can get the lower bound $u_d$. In conformal prediction, $c^*$ is the $1-\epsilon$ quantile of $S_{nc}$. Inversely, $\epsilon = 1 - \int_0^{c^*} f_{nc}(x) dx$. Hence, $u_d$ is bounded by $1-\epsilon = \int_0^{c^*} f_{nc}(x) dx$. Conservatively, we consider $u_d = \int_0^{c_*} f_{nc}(x) dx$.
\end{proof}

\paragraph{Justification on Separating Decision Uncertainty}
Formal verification can be computationally expensive and time-consuming when the number of atomic propositions grows. The complexity increases linearly with the number of states in the constructed Kripke structure and exponentially with the number of atomic propositions contained in the specifications \cite{model-checking}. Therefore, we do not directly verify the Kripke structures during planning. 
Instead, we obtain the calibration set and the nonconformity distribution offline, then estimate the decision uncertainty online. Such estimation (via Eq. \ref{eq: decision-unc-score}) is $O(1)$, which significantly reduces the complexity and turnaround time for planning.

In the planning stage, we estimate a decision uncertainty score $u_d$ and compare it against a predefined \emph{decision threshold} $t_d$. We only execute the plan if $u_d \ge t_d$. Hence, we can guarantee the probability of safe plan execution in the target environment. We demonstrate the procedure in Fig. \ref{fig: strategy}.

\section{From Uncertainty Disentanglement to Intervention}

Building on the uncertainty disentanglement, we next present two key interventions, \textit{Active Sensing} and \textit{Automated Refinement}, designed to reduce these uncertainties during model deployment. The active sensing mechanism leverages perception uncertainty scores to dynamically guide re-observations, ensuring propagation of high-confidence visual inputs for subsequent plan generation. Meanwhile, automated refinement systematically fine-tunes the model based on both perception and decision uncertainty, using high-certainty data to enhance the model’s ability to generate task-compliant instructions.
 
Together, these approaches create an iterative cycle, where active sensing feeds high-quality data into the planning module, and automated refinement iteratively improves the model's overall reliability, ultimately leading to a robust planning pipeline that effectively minimizes uncertainty from the perception to the decision-making stage.

\subsection{Active Sensing with Perception Uncertainty}
\label{sec: active-sensing}
We incorporate perception uncertainty scores into an active sensing process, using these scores as indicators to initiate re-observation when necessary. This approach allows assessment of whether the observed image meets a predefined certainty threshold, enhancing the reliability of perception outcomes and the subsequent text generation step.

Figure \ref{fig: strategy} illustrates the active sensing process. Given an image $I$ and a text description $P_T$, the vision encoder $V$ first extracts image embeddings. Following the steps in Section \ref{sec: conf-perc-unc-calibrate}, we obtain a perception uncertainty score $u_p$. If $u_p$ exceeds a specified \emph{perception threshold} $t_p \in [0, 1]$, we proceed to text generation. Otherwise, the robot is instructed to re-observe the scene by collecting a new image through camera rotation, switching to an alternate camera view, or by recapturing another image after a short delay until the perception uncertainty score rises above $t_p$. This iterative process continues until the perception uncertainty score surpasses the threshold. This active sensing mechanism ensures that only images with low perception uncertainty are fed into the text generator, reducing the risk of generating plans that fail due to perceptual inaccuracies.

\begin{algorithm}[t]
  \caption{Procedure for Automated Refinement}\label{alg: fine-tune}
  \begin{algorithmic}
    \STATE\algorithmicinput{ task bank $B_T$, image set $S_I$, foundation model $M$, atomic propositions set $AP$, specification set $\Phi$, sample size $N_S$, perception threshold $t_p$}
    \STATE\algorithmicoutput{ foundation model $M$}

    \STATE data = [] \hfill\COMMENT{Initialize an empty fine-tuning data set}

    \STATE\algorithmicwhile{ True }
        \STATE $\quad I \sim S_I, P_T \sim B_T$ \hfill\COMMENT{Sample $I$ and $P_T$}
        \STATE $\quad$ Obtain perception uncertainty score $u_p$ for $I$
        \STATE $\quad$\algorithmicif{ $u_p < t_p$}
        \STATE $\quad \quad$ \textbf{continue} \hfill\COMMENT{Skip high-uncertainty images}
        \STATE $\quad T, c = M (P_T, I)$ \hfill\COMMENT{Query foundation model}
        \STATE $\quad$Extract set of objects $Y$ from $I$
        \STATE $\quad \mathcal{A} = \textbf{pEnc}(T, AP, Y)$
        \STATE $\quad$ \algorithmicif{ $\forall_{\phi \in \Phi} \ \textbf{pEnc}(T, AP, Y) \models \phi$}
            \STATE $\quad \quad$ data.add($(I, P_T, T)$)
            \STATE $\quad \quad$ \algorithmicif{ size(data) $\ge N_S$}
                \STATE $\quad\quad\quad$ \textbf{break}
    \STATE Fine-tune $M$ with the collected data
  \end{algorithmic}
\end{algorithm}

\subsection{Automated Refinement Using Both Uncertainties}
\label{sec: refinement}
However, even with precise visual inputs, foundation models may still produce instructions that fail to meet task specifications due to a lack of domain knowledge. To address this, we introduce an automated refinement process aimed at minimizing the likelihood of generating such specification-violating instructions.

Consider an \emph{image set} containing task environment images and a \emph{task bank} of text-based task descriptions. This data can be collected from high-fidelity simulations or real task executions. The proposed automated refinement procedure is outlined in Algorithm \ref{alg: fine-tune}. In this process, we filter out images with high perception uncertainty, use the foundation model to generate instructions for each image-task description pair, verify the instructions against task specifications, and add only the valid instructions to the fine-tuning dataset.

The model is then fine-tuned in a supervised manner, where each image-task description pair serves as input, and the instruction is the expected output. If supervised fine-tuning is unavailable, we can generate positive and negative instruction pairs and apply direct preference optimization (DPO) \cite{rafailov2024direct}. Positive instructions are those whose automaton representations satisfy all specifications. Figure \ref{fig: fine-tune-pipeline} visually demonstrates this fine-tuning framework. This approach eliminates the need for human annotations or labels, allowing the model to be refined with virtually unlimited data samples until it reaches the desired performance.

\textbf{Note:} The refinement process occurs before grounding the planning pipeline in the task environment. Thus, we apply the FMDP described in Section \ref{sec: fmdp} to obtain a nonconformity distribution after refinement, which is then used to estimate decision uncertainty scores during planning.

\section{Experiments} \label{sec: results}

We perform simulated and real-world autonomous driving experiments to demonstrate \textbf{(1)} how we disentangle and quantify perception and decision uncertainty in driving tasks; \textbf{(2)} how the proposed active sensing mechanism enhances visual input quality and reduces perceptual errors; and \textbf{(3)} how the proposed uncertainty-aware automated fine-tuning framework refines the model to improve its capability to meet task specifications. In our experiments, we use LLaVA-1.5-7B \citep{liu2023improvedllava} for perception and plan generation; it consists of a CLIP-L-336px vision encoder \citep{clip} and a Vicuna-7B-v1.5 language generator \citep{zheng2023judging}.

\begin{figure}[H]
    \centering
    \resizebox{0.5\linewidth}{0.35\linewidth}{
\begin{tikzpicture}

\definecolor{darkgray176}{RGB}{176,176,176}
\definecolor{purple}{RGB}{128,0,128}

\begin{axis}[
tick align=outside,
tick pos=left,
title={\Large Nonconformity Scores (Perception)},
x grid style={darkgray176},
xmin=-0.02, xmax=1,
xtick style={color=black},
y grid style={darkgray176},
ylabel style={align=center},
ylabel={Proportion},
ymin=0, ymax=0.1,
ytick style={color=black},
label style={font=\Large},
yticklabel style={
  /pgf/number format/precision=3,
  /pgf/number format/fixed}
]
\draw[draw=black,fill=purple,fill opacity=0.75,line width=0.259723636363636pt] (axis cs:0.000202139800000001,0) rectangle (axis cs:0.019708564004,0.085);
\draw[draw=black,fill=purple,fill opacity=0.75,line width=0.259723636363636pt] (axis cs:0.019708564004,0) rectangle (axis cs:0.039214988208,0.06);
\draw[draw=black,fill=purple,fill opacity=0.75,line width=0.259723636363636pt] (axis cs:0.039214988208,0) rectangle (axis cs:0.058721412412,0.05);
\draw[draw=black,fill=purple,fill opacity=0.75,line width=0.259723636363636pt] (axis cs:0.058721412412,0) rectangle (axis cs:0.078227836616,0.02);
\draw[draw=black,fill=purple,fill opacity=0.75,line width=0.259723636363636pt] (axis cs:0.078227836616,0) rectangle (axis cs:0.09773426082,0.03);
\draw[draw=black,fill=purple,fill opacity=0.75,line width=0.259723636363636pt] (axis cs:0.09773426082,0) rectangle (axis cs:0.117240685024,0.02);
\draw[draw=black,fill=purple,fill opacity=0.75,line width=0.259723636363636pt] (axis cs:0.117240685024,0) rectangle (axis cs:0.136747109228,0.03);
\draw[draw=black,fill=purple,fill opacity=0.75,line width=0.259723636363636pt] (axis cs:0.136747109228,0) rectangle (axis cs:0.156253533432,0.035);
\draw[draw=black,fill=purple,fill opacity=0.75,line width=0.259723636363636pt] (axis cs:0.156253533432,0) rectangle (axis cs:0.175759957636,0.015);
\draw[draw=black,fill=purple,fill opacity=0.75,line width=0.259723636363636pt] (axis cs:0.175759957636,0) rectangle (axis cs:0.19526638184,0.02);
\draw[draw=black,fill=purple,fill opacity=0.75,line width=0.259723636363636pt] (axis cs:0.19526638184,0) rectangle (axis cs:0.214772806044,0.01);
\draw[draw=black,fill=purple,fill opacity=0.75,line width=0.259723636363636pt] (axis cs:0.214772806044,0) rectangle (axis cs:0.234279230248,0.01);
\draw[draw=black,fill=purple,fill opacity=0.75,line width=0.259723636363636pt] (axis cs:0.234279230248,0) rectangle (axis cs:0.253785654452,0.01);
\draw[draw=black,fill=purple,fill opacity=0.75,line width=0.259723636363636pt] (axis cs:0.253785654452,0) rectangle (axis cs:0.273292078656,0.015);
\draw[draw=black,fill=purple,fill opacity=0.75,line width=0.259723636363636pt] (axis cs:0.273292078656,0) rectangle (axis cs:0.29279850286,0.015);
\draw[draw=black,fill=purple,fill opacity=0.75,line width=0.259723636363636pt] (axis cs:0.29279850286,0) rectangle (axis cs:0.312304927064,0.01);
\draw[draw=black,fill=purple,fill opacity=0.75,line width=0.259723636363636pt] (axis cs:0.312304927064,0) rectangle (axis cs:0.331811351268,0.025);
\draw[draw=black,fill=purple,fill opacity=0.75,line width=0.259723636363636pt] (axis cs:0.331811351268,0) rectangle (axis cs:0.351317775472,0.01);
\draw[draw=black,fill=purple,fill opacity=0.75,line width=0.259723636363636pt] (axis cs:0.351317775472,0) rectangle (axis cs:0.370824199676,0.015);
\draw[draw=black,fill=purple,fill opacity=0.75,line width=0.259723636363636pt] (axis cs:0.370824199676,0) rectangle (axis cs:0.39033062388,0.01);
\draw[draw=black,fill=purple,fill opacity=0.75,line width=0.259723636363636pt] (axis cs:0.39033062388,0) rectangle (axis cs:0.409837048084,0.005);
\draw[draw=black,fill=purple,fill opacity=0.75,line width=0.259723636363636pt] (axis cs:0.409837048084,0) rectangle (axis cs:0.429343472288,0.025);
\draw[draw=black,fill=purple,fill opacity=0.75,line width=0.259723636363636pt] (axis cs:0.429343472288,0) rectangle (axis cs:0.448849896492,0.025);
\draw[draw=black,fill=purple,fill opacity=0.75,line width=0.259723636363636pt] (axis cs:0.448849896492,0) rectangle (axis cs:0.468356320696,0.005);
\draw[draw=black,fill=purple,fill opacity=0.75,line width=0.259723636363636pt] (axis cs:0.468356320696,0) rectangle (axis cs:0.4878627449,0.02);
\draw[draw=black,fill=purple,fill opacity=0.75,line width=0.259723636363636pt] (axis cs:0.4878627449,0) rectangle (axis cs:0.507369169104,0.005);
\draw[draw=black,fill=purple,fill opacity=0.75,line width=0.259723636363636pt] (axis cs:0.507369169104,0) rectangle (axis cs:0.526875593308,0.005);
\draw[draw=black,fill=purple,fill opacity=0.75,line width=0.259723636363636pt] (axis cs:0.526875593308,0) rectangle (axis cs:0.546382017512,0);
\draw[draw=black,fill=purple,fill opacity=0.75,line width=0.259723636363636pt] (axis cs:0.546382017512,0) rectangle (axis cs:0.565888441716,0.01);
\draw[draw=black,fill=purple,fill opacity=0.75,line width=0.259723636363636pt] (axis cs:0.565888441716,0) rectangle (axis cs:0.58539486592,0.005);
\draw[draw=black,fill=purple,fill opacity=0.75,line width=0.259723636363636pt] (axis cs:0.58539486592,0) rectangle (axis cs:0.604901290124,0.005);
\draw[draw=black,fill=purple,fill opacity=0.75,line width=0.259723636363636pt] (axis cs:0.604901290124,0) rectangle (axis cs:0.624407714328,0.01);
\draw[draw=black,fill=purple,fill opacity=0.75,line width=0.259723636363636pt] (axis cs:0.624407714328,0) rectangle (axis cs:0.643914138532,0.025);
\draw[draw=black,fill=purple,fill opacity=0.75,line width=0.259723636363636pt] (axis cs:0.643914138532,0) rectangle (axis cs:0.663420562736,0.015);
\draw[draw=black,fill=purple,fill opacity=0.75,line width=0.259723636363636pt] (axis cs:0.663420562736,0) rectangle (axis cs:0.68292698694,0.015);
\draw[draw=black,fill=purple,fill opacity=0.75,line width=0.259723636363636pt] (axis cs:0.68292698694,0) rectangle (axis cs:0.702433411144,0.035);
\draw[draw=black,fill=purple,fill opacity=0.75,line width=0.259723636363636pt] (axis cs:0.702433411144,0) rectangle (axis cs:0.721939835348,0.03);
\draw[draw=black,fill=purple,fill opacity=0.75,line width=0.259723636363636pt] (axis cs:0.721939835348,0) rectangle (axis cs:0.741446259552,0.025);
\draw[draw=black,fill=purple,fill opacity=0.75,line width=0.259723636363636pt] (axis cs:0.741446259552,0) rectangle (axis cs:0.760952683756,0.005);
\draw[draw=black,fill=purple,fill opacity=0.75,line width=0.259723636363636pt] (axis cs:0.760952683756,0) rectangle (axis cs:0.78045910796,0.025);
\draw[draw=black,fill=purple,fill opacity=0.75,line width=0.259723636363636pt] (axis cs:0.78045910796,0) rectangle (axis cs:0.799965532164,0.025);
\draw[draw=black,fill=purple,fill opacity=0.75,line width=0.259723636363636pt] (axis cs:0.799965532164,0) rectangle (axis cs:0.819471956368,0.015);
\draw[draw=black,fill=purple,fill opacity=0.75,line width=0.259723636363636pt] (axis cs:0.819471956368,0) rectangle (axis cs:0.838978380572,0.01);
\draw[draw=black,fill=purple,fill opacity=0.75,line width=0.259723636363636pt] (axis cs:0.838978380572,0) rectangle (axis cs:0.858484804776,0.03);
\draw[draw=black,fill=purple,fill opacity=0.75,line width=0.259723636363636pt] (axis cs:0.858484804776,0) rectangle (axis cs:0.87799122898,0.03);
\draw[draw=black,fill=purple,fill opacity=0.75,line width=0.259723636363636pt] (axis cs:0.87799122898,0) rectangle (axis cs:0.897497653184,0.02);
\draw[draw=black,fill=purple,fill opacity=0.75,line width=0.259723636363636pt] (axis cs:0.897497653184,0) rectangle (axis cs:0.917004077388,0.045);
\draw[draw=black,fill=purple,fill opacity=0.75,line width=0.259723636363636pt] (axis cs:0.917004077388,0) rectangle (axis cs:0.936510501592,0.01);
\draw[draw=black,fill=purple,fill opacity=0.75,line width=0.259723636363636pt] (axis cs:0.936510501592,0) rectangle (axis cs:0.956016925796,0.015);
\draw[draw=black,fill=purple,fill opacity=0.75,line width=0.259723636363636pt] (axis cs:0.956016925796,0) rectangle (axis cs:0.97552335,0.01);
\end{axis}

\end{tikzpicture}}
    \resizebox{0.45\linewidth}{0.35\linewidth}{
\begin{tikzpicture}

\definecolor{darkgray176}{RGB}{176,176,176}
\definecolor{steelblue31119180}{RGB}{31,119,180}

\begin{axis}[
tick align=outside,
tick pos=left,
title={\Large Nonconformity Scores (Decision)},
x grid style={darkgray176},
xmin=-0.02, xmax=1.02,
xtick style={color=black},
y grid style={darkgray176},
ymin=0, ymax=0.1,
ytick style={color=black},
yticklabel style={
  /pgf/number format/precision=3,
  /pgf/number format/fixed}
]
\draw[draw=black,fill=steelblue31119180,fill opacity=0.75,line width=0.259723636363636pt] (axis cs:8.71432922090049e-07,0) rectangle (axis cs:0.0200008540042637,0.0574314574314574);
\draw[draw=black,fill=steelblue31119180,fill opacity=0.75,line width=0.259723636363636pt] (axis cs:0.0200008540042637,0) rectangle (axis cs:0.0400008365756052,0.0173160173160173);
\draw[draw=black,fill=steelblue31119180,fill opacity=0.75,line width=0.259723636363636pt] (axis cs:0.0400008365756052,0) rectangle (axis cs:0.0600008191469468,0.0178932178932179);
\draw[draw=black,fill=steelblue31119180,fill opacity=0.75,line width=0.259723636363636pt] (axis cs:0.0600008191469468,0) rectangle (axis cs:0.0800008017182883,0.0152958152958153);
\draw[draw=black,fill=steelblue31119180,fill opacity=0.75,line width=0.259723636363636pt] (axis cs:0.0800008017182883,0) rectangle (axis cs:0.10000078428963,0.010966810966811);
\draw[draw=black,fill=steelblue31119180,fill opacity=0.75,line width=0.259723636363636pt] (axis cs:0.10000078428963,0) rectangle (axis cs:0.120000766860971,0.0121212121212121);
\draw[draw=black,fill=steelblue31119180,fill opacity=0.75,line width=0.259723636363636pt] (axis cs:0.120000766860971,0) rectangle (axis cs:0.140000749432313,0.0147186147186147);
\draw[draw=black,fill=steelblue31119180,fill opacity=0.75,line width=0.259723636363636pt] (axis cs:0.140000749432313,0) rectangle (axis cs:0.160000732003655,0.012987012987013);
\draw[draw=black,fill=steelblue31119180,fill opacity=0.75,line width=0.259723636363636pt] (axis cs:0.160000732003655,0) rectangle (axis cs:0.180000714574996,0.012987012987013);
\draw[draw=black,fill=steelblue31119180,fill opacity=0.75,line width=0.259723636363636pt] (axis cs:0.180000714574996,0) rectangle (axis cs:0.200000697146338,0.0135642135642136);
\draw[draw=black,fill=steelblue31119180,fill opacity=0.75,line width=0.259723636363636pt] (axis cs:0.200000697146338,0) rectangle (axis cs:0.220000679717679,0.0135642135642136);
\draw[draw=black,fill=steelblue31119180,fill opacity=0.75,line width=0.259723636363636pt] (axis cs:0.220000679717679,0) rectangle (axis cs:0.240000662289021,0.0164502164502165);
\draw[draw=black,fill=steelblue31119180,fill opacity=0.75,line width=0.259723636363636pt] (axis cs:0.240000662289021,0) rectangle (axis cs:0.260000644860362,0.0193362193362193);
\draw[draw=black,fill=steelblue31119180,fill opacity=0.75,line width=0.259723636363636pt] (axis cs:0.260000644860362,0) rectangle (axis cs:0.280000627431704,0.0196248196248196);
\draw[draw=black,fill=steelblue31119180,fill opacity=0.75,line width=0.259723636363636pt] (axis cs:0.280000627431704,0) rectangle (axis cs:0.300000610003046,0.0173160173160173);
\draw[draw=black,fill=steelblue31119180,fill opacity=0.75,line width=0.259723636363636pt] (axis cs:0.300000610003046,0) rectangle (axis cs:0.320000592574387,0.0173160173160173);
\draw[draw=black,fill=steelblue31119180,fill opacity=0.75,line width=0.259723636363636pt] (axis cs:0.320000592574387,0) rectangle (axis cs:0.340000575145729,0.0210678210678211);
\draw[draw=black,fill=steelblue31119180,fill opacity=0.75,line width=0.259723636363636pt] (axis cs:0.340000575145729,0) rectangle (axis cs:0.36000055771707,0.025974025974026);
\draw[draw=black,fill=steelblue31119180,fill opacity=0.75,line width=0.259723636363636pt] (axis cs:0.36000055771707,0) rectangle (axis cs:0.380000540288412,0.023953823953824);
\draw[draw=black,fill=steelblue31119180,fill opacity=0.75,line width=0.259723636363636pt] (axis cs:0.380000540288412,0) rectangle (axis cs:0.400000522859753,0.025974025974026);
\draw[draw=black,fill=steelblue31119180,fill opacity=0.75,line width=0.259723636363636pt] (axis cs:0.400000522859753,0) rectangle (axis cs:0.420000505431095,0.0248196248196248);
\draw[draw=black,fill=steelblue31119180,fill opacity=0.75,line width=0.259723636363636pt] (axis cs:0.420000505431095,0) rectangle (axis cs:0.440000488002436,0.0329004329004329);
\draw[draw=black,fill=steelblue31119180,fill opacity=0.75,line width=0.259723636363636pt] (axis cs:0.440000488002436,0) rectangle (axis cs:0.460000470573778,0.032034632034632);
\draw[draw=black,fill=steelblue31119180,fill opacity=0.75,line width=0.259723636363636pt] (axis cs:0.460000470573778,0) rectangle (axis cs:0.48000045314512,0.0314574314574315);
\draw[draw=black,fill=steelblue31119180,fill opacity=0.75,line width=0.259723636363636pt] (axis cs:0.48000045314512,0) rectangle (axis cs:0.500000435716461,0.0331890331890332);
\draw[draw=black,fill=steelblue31119180,fill opacity=0.75,line width=0.259723636363636pt] (axis cs:0.500000435716461,0) rectangle (axis cs:0.520000418287803,0.0331890331890332);
\draw[draw=black,fill=steelblue31119180,fill opacity=0.75,line width=0.259723636363636pt] (axis cs:0.520000418287803,0) rectangle (axis cs:0.540000400859144,0.0288600288600289);
\draw[draw=black,fill=steelblue31119180,fill opacity=0.75,line width=0.259723636363636pt] (axis cs:0.540000400859144,0) rectangle (axis cs:0.560000383430486,0.0401154401154401);
\draw[draw=black,fill=steelblue31119180,fill opacity=0.75,line width=0.259723636363636pt] (axis cs:0.560000383430486,0) rectangle (axis cs:0.580000366001827,0.0317460317460317);
\draw[draw=black,fill=steelblue31119180,fill opacity=0.75,line width=0.259723636363636pt] (axis cs:0.580000366001827,0) rectangle (axis cs:0.600000348573169,0.0360750360750361);
\draw[draw=black,fill=steelblue31119180,fill opacity=0.75,line width=0.259723636363636pt] (axis cs:0.600000348573169,0) rectangle (axis cs:0.62000033114451,0.0303030303030303);
\draw[draw=black,fill=steelblue31119180,fill opacity=0.75,line width=0.259723636363636pt] (axis cs:0.62000033114451,0) rectangle (axis cs:0.640000313715852,0.0308802308802309);
\draw[draw=black,fill=steelblue31119180,fill opacity=0.75,line width=0.259723636363636pt] (axis cs:0.640000313715852,0) rectangle (axis cs:0.660000296287194,0.0314574314574315);
\draw[draw=black,fill=steelblue31119180,fill opacity=0.75,line width=0.259723636363636pt] (axis cs:0.660000296287194,0) rectangle (axis cs:0.680000278858535,0.0256854256854257);
\draw[draw=black,fill=steelblue31119180,fill opacity=0.75,line width=0.259723636363636pt] (axis cs:0.680000278858535,0) rectangle (axis cs:0.700000261429877,0.0173160173160173);
\draw[draw=black,fill=steelblue31119180,fill opacity=0.75,line width=0.259723636363636pt] (axis cs:0.700000261429877,0) rectangle (axis cs:0.720000244001218,0.0167388167388167);
\draw[draw=black,fill=steelblue31119180,fill opacity=0.75,line width=0.259723636363636pt] (axis cs:0.720000244001218,0) rectangle (axis cs:0.74000022657256,0.0152958152958153);
\draw[draw=black,fill=steelblue31119180,fill opacity=0.75,line width=0.259723636363636pt] (axis cs:0.74000022657256,0) rectangle (axis cs:0.760000209143901,0.00981240981240981);
\draw[draw=black,fill=steelblue31119180,fill opacity=0.75,line width=0.259723636363636pt] (axis cs:0.760000209143901,0) rectangle (axis cs:0.780000191715243,0.00923520923520924);
\draw[draw=black,fill=steelblue31119180,fill opacity=0.75,line width=0.259723636363636pt] (axis cs:0.780000191715243,0) rectangle (axis cs:0.800000174286584,0.0115440115440115);
\draw[draw=black,fill=steelblue31119180,fill opacity=0.75,line width=0.259723636363636pt] (axis cs:0.800000174286585,0) rectangle (axis cs:0.820000156857926,0.0132756132756133);
\draw[draw=black,fill=steelblue31119180,fill opacity=0.75,line width=0.259723636363636pt] (axis cs:0.820000156857926,0) rectangle (axis cs:0.840000139429268,0.00173160173160173);
\draw[draw=black,fill=steelblue31119180,fill opacity=0.75,line width=0.259723636363636pt] (axis cs:0.840000139429268,0) rectangle (axis cs:0.860000122000609,0.00317460317460317);
\draw[draw=black,fill=steelblue31119180,fill opacity=0.75,line width=0.259723636363636pt] (axis cs:0.860000122000609,0) rectangle (axis cs:0.880000104571951,0.00692640692640693);
\draw[draw=black,fill=steelblue31119180,fill opacity=0.75,line width=0.259723636363636pt] (axis cs:0.880000104571951,0) rectangle (axis cs:0.900000087143292,0.00606060606060606);
\draw[draw=black,fill=steelblue31119180,fill opacity=0.75,line width=0.259723636363636pt] (axis cs:0.900000087143292,0) rectangle (axis cs:0.920000069714634,0.00663780663780664);
\draw[draw=black,fill=steelblue31119180,fill opacity=0.75,line width=0.259723636363636pt] (axis cs:0.920000069714634,0) rectangle (axis cs:0.940000052285975,0.00923520923520924);
\draw[draw=black,fill=steelblue31119180,fill opacity=0.75,line width=0.259723636363636pt] (axis cs:0.940000052285975,0) rectangle (axis cs:0.960000034857317,0.00490620490620491);
\draw[draw=black,fill=steelblue31119180,fill opacity=0.75,line width=0.259723636363636pt] (axis cs:0.960000034857317,0) rectangle (axis cs:0.980000017428658,0.00634920634920635);
\draw[draw=black,fill=steelblue31119180,fill opacity=0.75,line width=0.259723636363636pt] (axis cs:0.980000017428658,0) rectangle (axis cs:1,0.0331890331890332);
\end{axis}

\end{tikzpicture}}
    \caption{The first and second figures from left depict nonconformity distributions for perception and plan generation in simulated driving scenes (Carla) respectively.}
    \label{fig: unc-dist}
    \vspace{-10pt}
\end{figure}
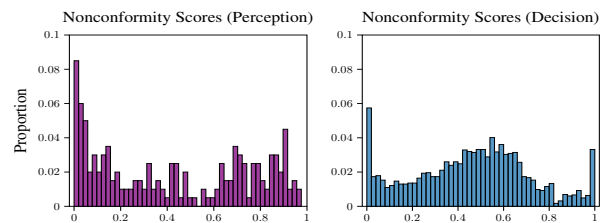

\subsection{Uncertainty Disentanglement and Quantification}
\label{sec: experiment-quant}
\textbf{Perception Uncertainty:}
We collect 542 images with ground truth labels from the Carla simulator, forming a \textit{calibration set}. Then, we apply CLIP with a pre-trained projection head to classify objects such as stop signs, cars, and pedestrians. We classify a total of 3000 objects within the 542 images and collect their confidence scores. Then, we obtain a nonconformity distribution, as shown in Fig. \ref{fig: unc-dist}, following the procedure outlined in Sec. \ref{sec: conf-perc-unc-calibrate}. 

\begin{figure*}[t]
    \centering
    \includegraphics[width=0.9\linewidth]{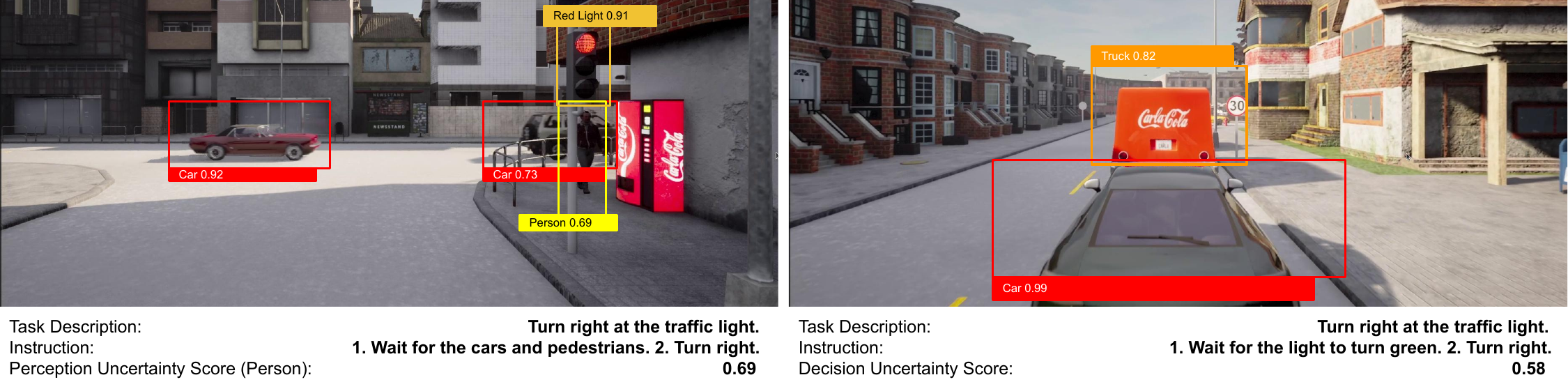}
    \caption{The first figure shows a scenario with high perception uncertainty due to the shadow and occlusion (perception uncertainty scores next to bounding boxes). The second figure shows a scenario with low perception uncertainty but high decision uncertainty, due to the inconsistency between the image and task description (traffic light is absent).}
    \label{fig: uncertainty-case-study}
    \vspace{-10pt}
\end{figure*}

\textbf{Decision Uncertainty:}
We collect a \textit{calibration set} containing 400 images from the Carla simulator---each associated with a task description---to obtain another nonconformity distribution (Fig. \ref{fig: unc-dist}) using the procedure in Sec. \ref{sec: fmdp} and \ref{sec: estimate-dec-unc}. 

For each image, we randomly select a task description from a set of tasks \emph{$\{$go straight, turn left, turn right, make a U-turn$\}$ at the $\{$traffic light, stop sign$\}$}.
We define 10 safety-critical logical specifications for the driving tasks, some of which are listed below: 

$\phi_1 = \lalways ( \text{red light} \rightarrow \neg \text{move forward} )$,

$\phi_2 = \lalways ( \text{pedestrian} \rightarrow \text{wait} )$,

$\phi_3 = \lalways ( \neg \text{stop sign} \land \neg \text{ traffic light } \rightarrow \text{move forward} )$,

$\phi_4 = \lalways ( \text{green light} \land \neg \text{pedestrian} \rightarrow \lnext \neg \text{wait} )$,

$\phi_5 = \lalways ( (\text{stop sign} \land \neg \text{car} \land \neg \text{pedestrian}) \rightarrow \lnext \neg \text{wait} )$,

where $\lalways$, $\leventually$, and $\lnext$ refer to always, eventually, and next, respectively. The detailed task descriptions for the foundation model and the complete set of logical specifications are provided in Appendix \ref{app: prompt} and \ref{app: drive-specs} respectively.

To avoid scenarios where the generated plans fail to meet the specifications due to language ambiguity or mismatching synonyms, we include in-context examples to constrain the sentence structure and vocabulary of the generated plans.

\textbf{Quantification:} 
Fig. \ref{fig: unc-dist} shows the nonconformity distribution we obtained. We can estimate a probability density function for each distribution.
For every object detection and corresponding confidence score, we use Eq. \ref{eq: perception-unc-score} to calibrate the confidence into perception uncertainty scores. We present examples of the calibrated scores in Fig. \ref{fig: uncertainty-case-study} (next to the bounding boxes) for both types of uncertainty.

For every plan generated by the model, we follow the procedure to prompt the model at the beginning of Sec. \ref{sec: decision-unc} and obtain a confidence score and compute the decision uncertainty scores via Eq. \ref{eq: decision-unc-score}. 
We present a scenario with high perception uncertainty and one with high decision uncertainty in Fig. \ref{fig: uncertainty-case-study}. After uncertainty disentanglement and quantification, we can pinpoint the source of high uncertainty and tailor targeted interventions for mitigation.

\begin{figure*}[t]
    \centering
    \includegraphics[width=0.9\linewidth]{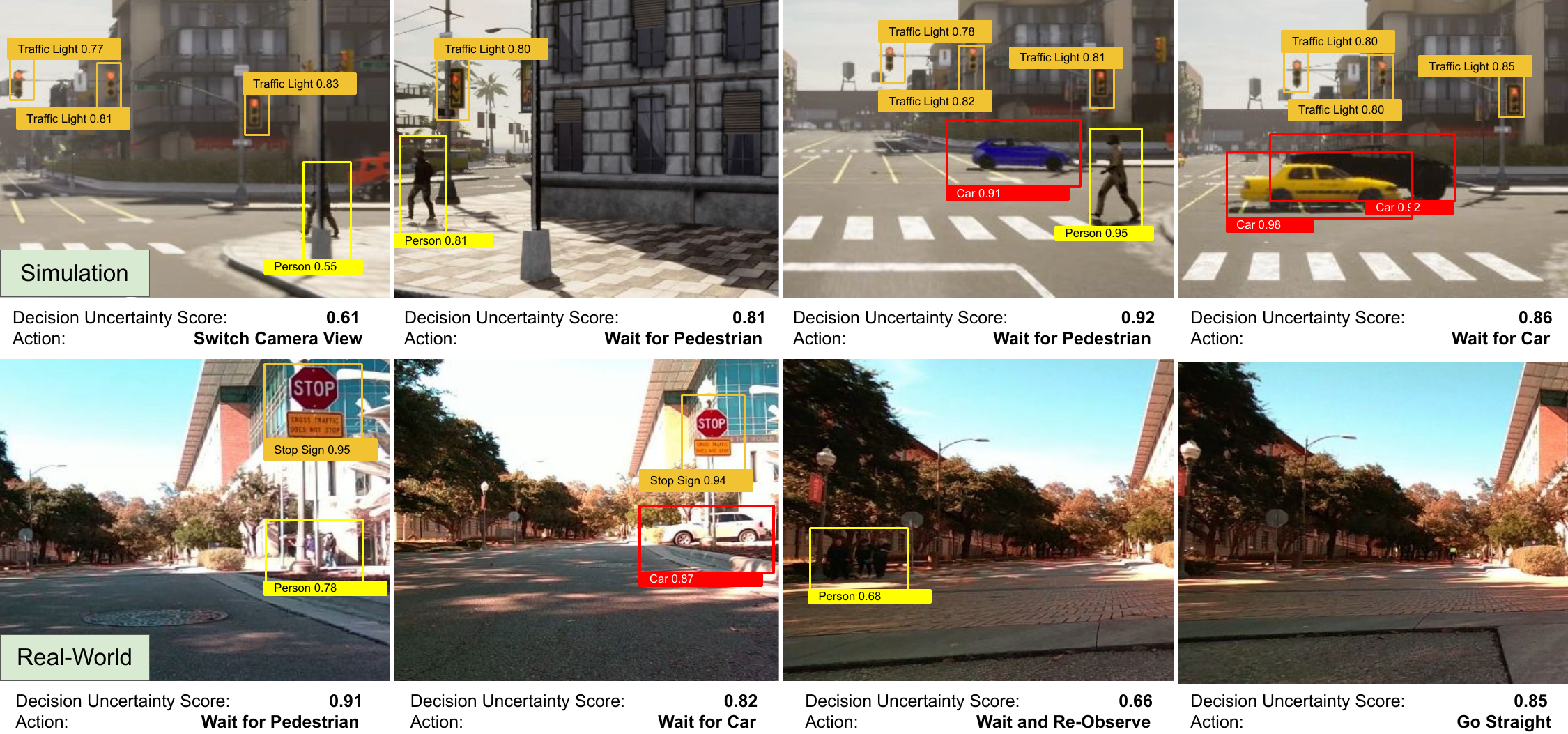}
    \caption{Illustration of autonomous driving tasks wherein our strategy presented in Fig. \ref{fig: strategy} satisfies all task specifications during plan execution in both simulations and real-world environments.}
    \label{fig: use-cases}
\end{figure*}

\vspace{-4pt}
\subsection{Targeted Intervention 1: Active Sensing}
\vspace{-4pt}
\label{sec: active-sensing-result}
We deploy the active sensing mechanism to the planning pipeline that uses a raw foundation model without fine-tuning, as described in Sec. \ref{sec: active-sensing}. We set the threshold $t_p = 0.7$, i.e., a $70\%$ percent probability (and above) of correct visual interpretation is acceptable. We also examine other thresholds and present results in the Appendix \ref{app: thresholds}.

On the simulated data, we observe a 35\% reduction in decision uncertainty. Across a \emph{test set} of 50 driving scenes from our ground robot and NuScenes data \cite{nuscene}, the active sensing mechanism reduces perception uncertainty by over 10\% and subsequently reduces decision uncertainty by 10\%. We present more results in Tab. \ref{tab: resutls}.

We present qualitative results in Fig. \ref{fig: use-cases}, illustrating successful plan executions with the active sensing mechanism in simulated and real-world autonomous driving scenes. 

\vspace{-4pt}
\subsection{Targeted Intervention 2: Automated Refinement}
\label{sec: refine-experiment}
\vspace{-4pt}
\begin{figure}[t]
    \centering
    \resizebox{0.7\linewidth}{0.5\linewidth}{\begin{tikzpicture}
\begin{axis}[
ylabel=Loss,
xlabel=Step,
xmin = 0,
xmax = 410,
legend pos=north east,
]

\addplot[orange] table [x=step, y=benchmark, col sep=comma] {figures/csv/loss.csv};
\addlegendentry{Benchmark}

\addplot[blue] table [x=step, y=ours, col sep=comma] {figures/csv/loss.csv};
\addlegendentry{Ours}

\end{axis}
\end{tikzpicture}}
    \caption{A comparison of the fine-tuning cross-entropy loss obtained using our framework and the benchmark.}
    \label{fig: loss}
    \vspace{-15pt}
\end{figure}
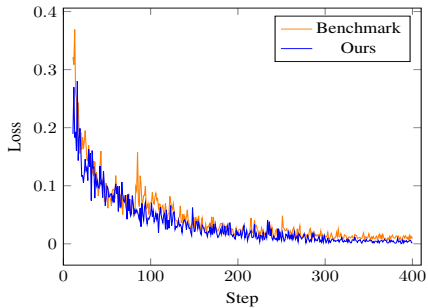

\begin{figure}[t]
    \centering
    \includegraphics[width=0.7\linewidth]{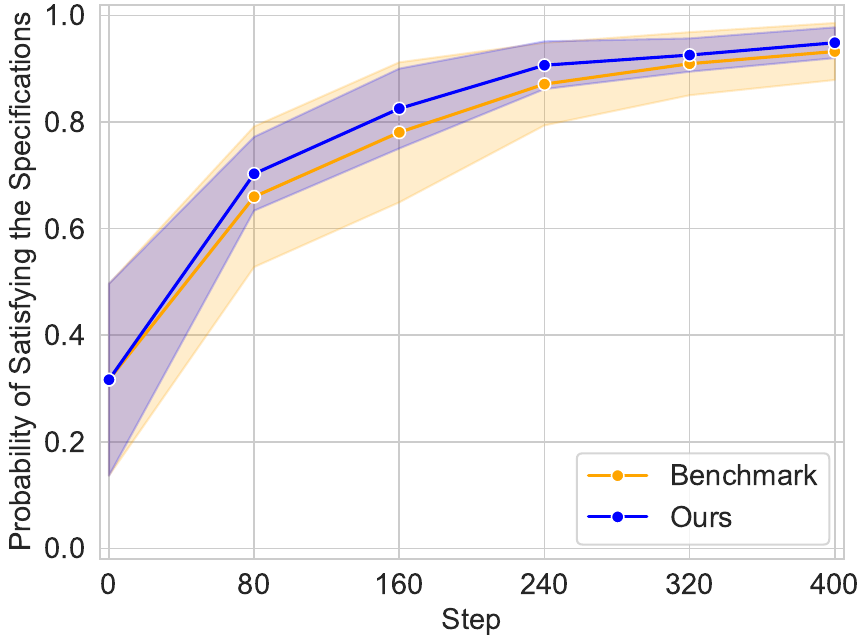}
    \caption{Probability of specification compliance during fine-tuning (color band shows standard deviation (SD)).}
    \label{fig: fine-tune-result}
    \vspace{-15pt}
\end{figure}

We investigate a complementary intervention that we cascade on top of active sensing, building upon its benefits. To improve the foundation model's capability to generate plans satisfying specifications, we apply our automated refinement framework to fine-tune the foundation model for autonomous driving tasks using a \textit{train set} containing $800$ images from the Carla simulator and the aforementioned 10 pre-defined logical specifications. We follow the steps outlined in Sec. \ref{sec: refinement} to automatically generate refinement data and fine-tune the foundation model's text generator.

\textbf{Benchmark: } To showcase the necessity of our uncertainty disentanglement framework, we select a refinement benchmark \cite{DBLP:conf/mlsys/YangBIWCWT24} that fine-tunes the same model using the dataset without uncertainty disentanglement.

This framework works as follows: (1) Create an empty set of fine-tuning data, (2) select an image with a task description and query the foundation model for a plan, (3) build automaton-based representation for the plan and verify it against the specifications, (4) add the plan to the set of fine-tuning data (without uncertainty disentanglement) if it passes all the specifications, (5) repeat step 1-5 until sufficient data is obtained, and (6) freeze the vision encoder and fine-tune the model.

In contrast to the benchmark, our framework first quantifies perception uncertainty and filters out high-uncertainty data, hence enhancing data quality.

\begin{table*}[t]
    \centering
    \caption{Real-world driving results show that active sensing (AS) and automated refinement improve average perception and decision uncertainty scores in addition to the average and standard deviation of the likelihood of specification compliance. The raw model refers to the pretrained foundation model without any fine-tuning.}
    \vspace{-4pt}
    \begin{tabular}{cccccc} \toprule
         Planning Pipeline & \shortstack{Avg. Percp. \\ Unc. Score}  & \shortstack{Avg. Dec. \\ Unc. Score} & \shortstack{Prob. of Satisfying \\ Spec. (Avg)} & \shortstack{Prob. of Satisfying \\ Spec. (SD)} \\ \midrule
         Raw Model w/o AS &  0.842 & 0.279 & --- & ---\\
         Raw Model with AS & \textbf{0.936} & 0.306 & 0.316 & 0.180 \\
         Fine-tuned Model (Benchmark) with AS & \textbf{0.936} & 0.931 & 0.933 & 0.048 \\
         Fine-tuned Model (Ours) with AS & \textbf{0.936} & \textbf{0.955} & \textbf{0.959} & \textbf{0.025} \\\bottomrule
    \end{tabular}
    \label{tab: resutls}
\end{table*}

\textbf{Experiment Setting:} We fine-tune two identical LLaVA models using our framework and the benchmark and present the training losses in Fig. \ref{fig: loss}. Our framework offers smoother convergence to a lower loss compared to the benchmark, indicating a higher potential for generating specification-compliant plans. During fine-tuning, we save a checkpoint every 80 steps. 
Next, we use the \emph{test set} with 50 scenes (same as Sec. \ref{sec: active-sensing-result}) to validate the fine-tuned models. We use each checkpoint to generate plans given those images with task descriptions and record the probability that the plans satisfy the specifications.

\textbf{Result and Analysis:} 
Fig. \ref{fig: fine-tune-result} and Tab. \ref{tab: resutls} show the improvement in the model's ability to generate specification-compliant plans. 

We observe over 50\% improvement in the probability of generated plans satisfying the specifications after fine-tuning. Meanwhile, our framework outperforms the benchmark by improving this probability by up to \textbf{5\%} while reducing the standard deviation by \textbf{40\%}. The results indicate enhanced robustness and stability offered by our refinement framework, lowering uncertainty during planning. 
Uncertain visual observations increase the likelihood of generating plans that violate the specifications. For example, in the first image in Fig \ref{fig: uncertainty-case-study}, uncertainty in observing the occluded pedestrian may result in a plan that overlooks the pedestrian, causing safety violations. Therefore, filtering out high-uncertainty images using our framework improves model performance consistency and scalability.

\textbf{Sim2Real Transfer:} Moreover, we fine-tune the model solely on simulated data and yet achieve over $95\%$ probability of satisfying the specifications on real-world data, indicating an effective Sim2Real transfer of our uncertainty disentanglement framework and interventions. We present theoretical and empirical backing for this Sim2Real transfer capability in Appendix \ref{app: distribution}.

\begin{figure}[t]
    \centering
    \includegraphics[width=0.49\linewidth]{figures/demo/example1.jpg}
    \includegraphics[width=0.49\linewidth]{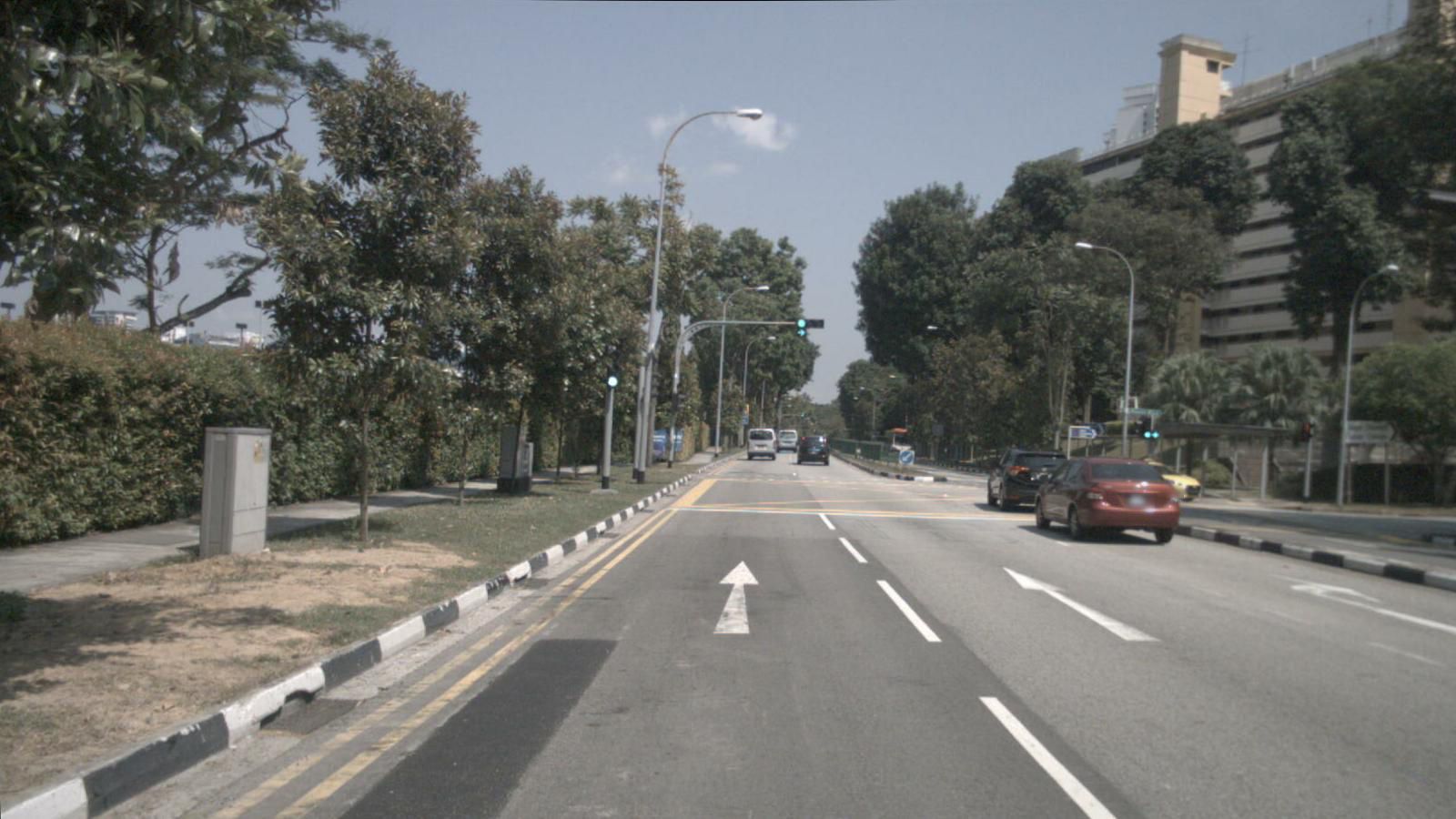}
    \caption{Two examples where the foundation model fine-tuned via our refinement framework passes all the specifications while the benchmark fails. The first figure shows a scene with a red light and pedestrians crossing. The second figure shows a green light ahead.}
    \label{fig: fine-tune-examples}
    \vspace{-15pt}
\end{figure}

\textbf{Case Study:} 
Fig. \ref{fig: fine-tune-examples} shows two examples indicating the superior performance of our refinement framework compared to the benchmark. In the first example, we feed the first image in Fig. \ref{fig: fine-tune-examples} along with a task description ``turn right at the stop sign'' to the foundation model and query for a plan. Note that this description is inconsistent with the image observation, as there is no stop sign in the image. We expect the foundation model to identify such inconsistency.

Prior to refinement, the raw model returns the following:
\vspace{4pt}
\begin{lstlisting}[language=completion]
    <completion>1. Pay attention to the stop sign.
    2. Turn right onto the street.</completion>
\end{lstlisting}
The raw model completely ignores the inconsistency and instructs to turn right directly, which violates $\phi_2$ in Sec. \ref{sec: experiment-quant} (always wait for pedestrians). This can lead to severe consequences and is a dangerous action to execute.

The fine-tuned model using the benchmark returns:
\vspace{4pt}
\begin{lstlisting}[language=completion]
    <completion>1. There is no stop sign, move forward.</completion>
\end{lstlisting}
\vspace{4pt}
Compared to the raw model, it notices the image-description inconsistency. However, the decision to move forward violates $\phi_2$ in Sec. \ref{sec: experiment-quant}. In contrast, the model fine-tuned using our framework successfully identifies the inconsistency and generates a plan that satisfies all the specifications:
\vspace{4pt}
\begin{lstlisting}[language=completion]
    <completion>1. Wait for the light to turn green.
    2. Turn right.</completion>
\end{lstlisting}

In the second example, we feed the second image in Fig. \ref{fig: fine-tune-examples} with a task description ``go straight at the traffic light'' to the foundation model. The model fine-tuned using the benchmark framework returns the following:
\vspace{4pt}
\begin{lstlisting}[language=completion]
    <completion>1. Wait for the light to turn green. 
    2. Move straight ahead.</completion>
\end{lstlisting}
It does not consider that the light is already green and decides to wait, violating $\phi_4$. In contrast, the model fine-tuned using our framework mitigates this issue:
\vspace{4pt}
\begin{lstlisting}[language=completion]
    <completion>1. The traffic light is green. 
    2. Move forward.</completion>
\end{lstlisting}
In both examples, we demonstrate how our refinement framework outperforms the benchmark.

\textbf{Decision Uncertainty in Plan Execution:}
The fine-tuned model does not achieve a 100\% specification-satisfaction rate. We employ the decision uncertainty score to determine whether to execute the generated plans, as described at the end of Sec. \ref{sec: estimate-dec-unc}. In the 50 scenes, 4 of the generated pans violate the specifications, and all of these plans have decision uncertainty scores below the pre-defined threshold $t_d=0.7$. Consequently, restricting the execution of plans with uncertainty scores above $t_d$ reduces the probability of specification violation from \textbf{8\%} to \textbf{0\%}, underscoring the importance of decision uncertainty during planning.

\section{Conclusion}

We presented a novel framework for enhancing multimodal foundation models in robotic planning. Our framework disentangles perception uncertainty in visual interpretation and decision uncertainty in plan generation. The proposed quantification methods, leverage conformal prediction for perception and FMDP for decision-making. The targeted interventions improve model robustness by employing active sensing and automated refinement. Empirical results from both real and simulated robotic tasks show that our framework reduces performance variability and improves task success rates, attesting to the value of distinct uncertainty interventions. Future work will extend this approach by considering additional uncertainty types and exploring broader interventions such as task description optimization, paving the way for robust and reliable autonomous systems.

\section*{Acknowledgements}
\vspace{-4pt}
This research was supported by the following: DARPA ANSR program under grant RTX CW2231110, DARPA TIAMAT program under grant HR0011-24-9-0431, and the Army Research Lab under grant ARL W911NF-23-S-0001.

\bibliography{ref}
\bibliographystyle{mlsys2025}


\newpage

\appendix
\onecolumn

\section{Additional Background}
\paragraph{Confidence Scores}
A deep learning model produces an output with a confidence score $c \in [0, 1]$. The resulting confidence score represents the likelihood of the output being ``correct'' with respect to the ground truth labels in the training data.
Foundation models such as GPT-series and LLaMA-series produce a distribution of outputs, sample one output from the distribution, and return the sampled output to users. Each output within the distribution is associated with a softmax confidence score indicating its probability of being sampled.

\section{Demonstration of Algorithm \ref{alg: text2automata}}
\label{sec: algo-demo}
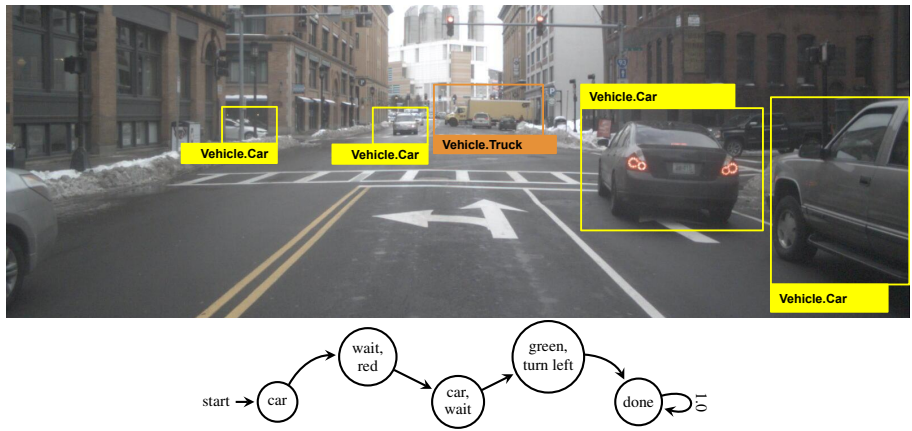
\begin{figure}[h]
    \centering
    \includegraphics[width=0.7\linewidth]{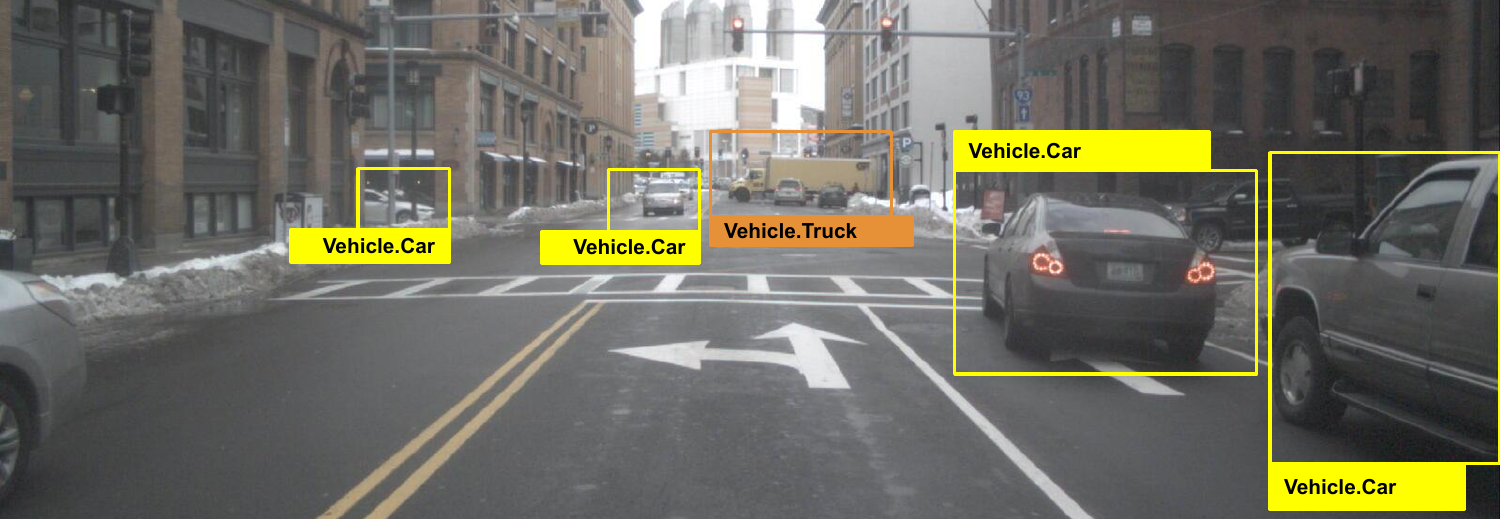}
    \begin{tikzpicture}[thick,scale=.6, node distance=2.2cm, every node/.style={transform shape}]
    \node[state, initial] (0) at (0, 0) {car};
    
    \node[state] (P1) at (2, 1) {\shortstack{wait, \\ red}};
    \node[state] (P1-1) at (6, 1) {\shortstack{green, \\ turn left}};
    
    \node[state] (P2) at (4, 0) {\shortstack{car, \\ wait}};

    \node[state] (done) at (8, 0) {done};
    
    \draw[->, shorten >=1pt, sloped, loop right] (done) to[left] node[above, align=center] {1.0} ();
    
    \draw[->, shorten >=1pt, sloped, bend left] (0) to[left] node[above, align=center] {} (P1);
    \draw[->, shorten >=1pt, sloped] (P1) to[left] node[above, align=center] {} (P2);
    
    \draw[->, shorten >=1pt, sloped] (P2) to[left] node[below, align=center] {} (P1-1);
    \draw[->, shorten >=1pt, sloped, bend left] (P1-1) to[left] node[above, align=center] {} (done);

\end{tikzpicture}
    \caption{A sample Kripke structure constructed from an image $I \in \mathcal{I}$ with observed objects $Y$ and a textual instruction $T$ using Algorithm \ref{alg: text2automata}.}
    \label{fig: text-aut-example}
\end{figure}

Given a set of atomic propositions $AP = \{$ green light, red light, car, pedestrian, wait, turn left,...$\}$. 
The robot agent sends an image as presented in Fig. \ref{fig: text-aut-example} with objects $Y = \{ \text{car, truck} \}$ and a task description ``turn left'' as inputs to the foundation model and obtains the instruction $T$:
\vspace{4pt}
\begin{lstlisting}[language=completion]
    <completion>1. Wait at the red light.
    2. Wait for opposite cars.
    3. Turn left at the green light.</completion>
\end{lstlisting}
The algorithm takes $T$, $Y$, and $AP$ as inputs and returns a Kripke structure, as presented in Fig. \ref{fig: text-aut-example}. In detail, the algorithm first parses $T$ into phrases
\vspace{4pt}
\begin{lstlisting}[language=completion]
    $Ph_1$ = <wait> <red light>
    $Ph_2$ = <wait> <car>
    $Ph_3$ = <turn left> <greed light>
\end{lstlisting}
Then, it creates an initial state with a label $Y = \{\text{car}\}$. Next, it adds one state per phrase, where the state label is the parsed verbs and nouns in the phrase. Last, it adds a final state and connects all the states sequentially. The resulting Kripke structure is in Fig. \ref{fig: text-aut-example}. 

We use a model checker named NuSMV \cite{Cimatti2002NuSMV} to verify the Kripke structure against the specifications. The Kripke structure in Fig. \ref{fig: text-aut-example} and the temporal logic specifications can be expressed as:
\vspace{13pt}
\begin{lstlisting}[language=NuSMV]
MODULE Plan_Kripke_Structure

VAR
  green_light : boolean;
  car : boolean;
  pedestrian : boolean;
  stop_sign : boolean;
  action : {wait, turn_left, turn_right, move_forward};

ASSIGN
  init(car) := TRUE;
  init(pedestrian) := FALSE;
  init(green_light) := FALSE;
  init(action) := wait;
  next(action) :=
    case
      green_traffic_light & !car : {turn_left};
      green_traffic_light & car : wait;
      !green_traffic_light : wait;
    esac;

LTLSPEC NAME sample_phi_1 :=
    G( !green_traffic_light -> ! action=move_forward );

LTLSPEC NAME sample_phi_2 :=
    G( pedestrian -> action=wait );
\end{lstlisting}

Next, we can run the following commands in the terminal to obtain the verification results:
\vspace{8pt}
\begin{lstlisting}[language=NuSMV]
#!NuSMV -source
read_model -i plan.smv # replace it with your file name
go

check_ltlspec -P "sample_phi_1" -o result1.txt 

check_ltlspec -P "sample_phi_2" -o result2.txt 

quit
\end{lstlisting}

Note that these commands will save the verification results into text files with the result in the following format:
\vspace{8pt}
\begin{lstlisting}[language=completion]
    -- specification  G ( !green_traffic_light -> ! action=move_forward)  is true
\end{lstlisting}

If the plan violates the specification, the result shows a counter-example, which is a sequence of states that violate the specification. For example,
\vspace{8pt}
\begin{lstlisting}[language=completion]
    -- specification  G ((car | pedestrian) ->  X action = wait)  is false
    -- as demonstrated by the following execution sequence
    Trace Description: LTL Counterexample 
    Trace Type: Counterexample 
      -> State: 1.1 <-
        pedestrian = FALSE
        car= FALSE
        stop_sign = FALSE
        action = move_forward
      -> State: 1.2 <-
        pedestrian = TRUE
      -> State: 1.3 <-
        pedestrian = FALSE
        action = turn_left
      -- Loop starts here
      -> State: 1.4 <-
        action = move_forward
      -> State: 1.5 <-
\end{lstlisting}

\section{Additional Experimental Details and Results}

\subsection{Task Description and Sample Response} \label{app: prompt}
During plan generation, we provide a set of examples to the foundation model to enforce its output structure. An example is presented below, where the blue texts represent the inputs to the foundation model, and the red texts represent the outputs generated from the foundation model.

\vspace{0.2cm}
\begin{lstlisting}[language=completion]
    <prompt>We: You are a robot driving a vehicle.

    User: <Image 1>
    User: Turn right at the traffic light.
    Assistant: 
    1. There is no traffic light visible, move forward.
    
    User: <Image 2>
    User: Turn right at the traffic light.
    Assistant: 
    1. Wait for light turn green.
    2. Turn right.
    
    User: <Image 3>
    User: Go straight at the traffic light.
    Assistant: 
    1. There is no traffic light visible, but there is a stop sign.
    2. Stop at the stop sign.
    3. Move forward.
    
    User: <Image 4>
    User: Turn left at the traffic light.
    Assistant: 
    1. Wait for light turn green.
    2. Wait for car.
    3. Move forward.
    
    User: <Image 5>
    User: Go straight at the stop sign.
    Assistant: 
    1. Stop at the stop sign.
    2. Wait for pedestrian.
    3. Move forward.
    
    User: <Image 6>
    User: Turn left at the stop sign.
    Assistant: 
    1. There is no stop sign visible, but there is a traffic light.
    2. Wait for light turn green.
    3. Turn left.
    
    User: <Image 7>
    User: Turn left at the stop sign.
    Assistant: 
    1. Stop at the stop sign.
    2. Wait for pedestrian.
    3. Turn left.
    
    User: <path to image>
    User: Turn right at the traffic light.
    You: </prompt><completion> 
    1. There is a traffic light ahead.
    2. Wait for the light to turn green.
    3. Turn right if there are no pedestrians.
    </completion>
    <prompt>We: does the plan satisfy the following rules:
        a. Yield for pedestrian
        b. Do not move forward or turn left at the red light
        
        Answer 'Y' or 'N' indicating yes or no.
    You: </prompt>
    <completion>Y</completion>
\end{lstlisting}
\vspace{0.2cm}
Note that the foundation model will return a single letter to indicate its choices.

\subsection{Autonomous Driving Task Specifications} 
\label{app: drive-specs}
To get the nonconformity distribution for decision uncertainty, we manually generate a set $\Phi$ of specifications regarding the driving rules. Next, we transform the textual decisions outputted from the foundation model to a Kripke structure through Algorithm \ref{alg: text2automata}. Then, we use a model checker to verify the Kripke structure against each specification $\phi \in \Phi$. In the experiment, the set of specifications $\Phi$ contains

$\phi_1 = \lalways ( \text{red light} \rightarrow \neg \text{move forward} )$,

$\phi_2 = \lalways ( \text{pedestrian} \rightarrow \text{wait} )$,

$\phi_3 = \lalways ( \neg \text{stop sign} \land \neg \text{ traffic light } \rightarrow \text{move forward} )$,

$\phi_4 = \lalways ( \text{green light} \land \neg \text{pedestrian} \rightarrow \lnext \neg \text{wait} )$,

$\phi_5 = \lalways ( (\text{stop sign} \land \neg \text{car} \land \neg \text{pedestrian}) \rightarrow \lnext \neg \text{wait} )$

$\phi_6 = \lalways (\text{opposite car} \rightarrow \neg \text{turn left})$,

$\phi_7 = \lalways ( \text{red light} \rightarrow \neg \text{turn left} )$,

$\phi_8 = \leventually \neg \text{wait} $,

$\Phi_9 = \lalways (\text{wait} \vee \text{move forward} \vee \text{turn left} \vee \text{turn right} )$,

$\Phi_{10} = \lalways ( \text{green light} \land \text{opposite car} \rightarrow ( \text{wait} \vee \text{move forward} \vee \text{turn right} ) )$.

\begin{figure}[t]
    \centering
    \resizebox{0.4\linewidth}{0.35\linewidth}{
\begin{tikzpicture}

\definecolor{darkgray176}{RGB}{176,176,176}
\definecolor{orange}{RGB}{255,165,0}
\definecolor{steelblue31119180}{RGB}{31,119,180}

\begin{axis}[
tick align=outside,
tick pos=left,
x grid style={darkgray176},
xlabel={CLIP Confidence Quantile in Real-World Images},
xmin=-0.05, xmax=1.05,
xtick style={color=black},
y grid style={darkgray176},
ylabel={CLIP Confidence Quantile in Carla Simulation},
ymin=-0.05, ymax=1.05,
ytick style={color=black}
]
\addplot [draw=white, fill=steelblue31119180, mark=*, only marks]
table{%
x  y
0 0
0.015 0
0.07 0.05
0.135 0.075
0.185 0.13
0.24 0.15
0.295 0.215
0.35 0.28
0.395 0.315
0.415 0.35
0.425 0.4
0.45 0.435
0.505 0.525
0.535 0.555
0.57 0.58
0.605 0.625
0.635 0.69
0.675 0.715
0.75 0.795
0.83 0.86
1 0.89
};
\addplot [semithick, orange, dashed]
table {%
0 0
1 1
};
\end{axis}

\end{tikzpicture}}
    \caption{Q-Q plot for the distributions of the calibration set from the Carla simulator and the images from the real-world driving environment. This plot shows that the vision encoder performs consistently on the calibration set and the real-world images, i.e., two datasets are identically distributed.}
    \label{fig: q-q}
\end{figure}
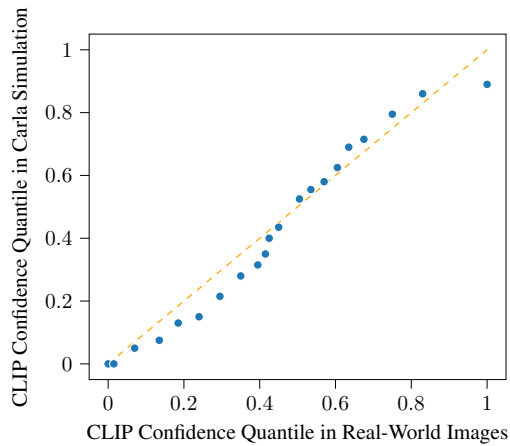

\subsection{Sim2Real Transfer}
\label{app: distribution}
Our uncertainty estimation framework assumes that the distribution of the calibration set is identically distributed with the test data distribution (\emph{Assumption 1}).

We empirically demonstrate that the data (images) collected from the Carla simulator and the images collected from the real world are identically distributed. Hence \emph{Assumption 1} holds. Therefore, the uncertainty score from the nonconformity distribution bounds the probability of correctly classifying objects in the environment.

We collect the confidence scores of the correct predictions from CLIP for the two sets of images and use a Q-Q plot that shows that the two sets of images are identically distributed.
The Q-Q plot \cite{gnanadesikan1968probability} compares the distribution of the confidence scores for the two datasets by plotting their quantiles against each other. The closer the scatter points are to the orange line ($y=x$), the higher the similarity between the two distributions.

This confirms that the assumption made for our Sim2Real transfer holds. Hence, we can apply the foundation model fine-tuned in the Carla simulator directly to the real-world driving environment.

The outcomes of both the fine-tuning experiments (Fig. \ref{fig: fine-tune-result}) and the planning demonstrations (Fig. \ref{fig: use-cases}) indicate that a foundation model, when fine-tuned and calibrated on simulation data, exhibit the capacity to be effectively transferred to real-world tasks without the need for domain-specific adjustments. 

\subsection{Out-of-Domain Application: Table-top Manipulation}
\label{app: arm-spec}

\begin{figure}[t]
    \centering
    \resizebox{0.4\linewidth}{0.35\linewidth}{
\begin{tikzpicture}[font=\normalfont]

\definecolor{darkgray176}{RGB}{176,176,176}
\definecolor{purple}{RGB}{128,0,128}

\begin{axis}[
tick align=outside,
tick pos=left,
title={Nonconformity Scores for Perception (Table-top)},
x grid style={darkgray176},
xlabel={Softmax},
xmin=-0.02, xmax=1,
xtick style={color=black},
y grid style={darkgray176},
ylabel style={align=center},
ylabel={Proportion},
ymin=0, ymax=0.1,
ytick style={color=black},
label style={font=\Large},
yticklabel style={
  /pgf/number format/precision=3,
  /pgf/number format/fixed}
]
\draw[draw=black,fill=purple,fill opacity=0.75,line width=0.259723636363636pt] (axis cs:0.0025939182347143,0) rectangle (axis cs:0.0221980951113455,0.065);
\draw[draw=black,fill=purple,fill opacity=0.75,line width=0.259723636363636pt] (axis cs:0.0221980951113455,0) rectangle (axis cs:0.0418022719879768,0.035);
\draw[draw=black,fill=purple,fill opacity=0.75,line width=0.259723636363636pt] (axis cs:0.0418022719879768,0) rectangle (axis cs:0.061406448864608,0.0275);
\draw[draw=black,fill=purple,fill opacity=0.75,line width=0.259723636363636pt] (axis cs:0.061406448864608,0) rectangle (axis cs:0.0810106257412393,0.045);
\draw[draw=black,fill=purple,fill opacity=0.75,line width=0.259723636363636pt] (axis cs:0.0810106257412393,0) rectangle (axis cs:0.100614802617871,0.055);
\draw[draw=black,fill=purple,fill opacity=0.75,line width=0.259723636363636pt] (axis cs:0.100614802617871,0) rectangle (axis cs:0.120218979494502,0.035);
\draw[draw=black,fill=purple,fill opacity=0.75,line width=0.259723636363636pt] (axis cs:0.120218979494502,0) rectangle (axis cs:0.139823156371133,0.04);
\draw[draw=black,fill=purple,fill opacity=0.75,line width=0.259723636363636pt] (axis cs:0.139823156371133,0) rectangle (axis cs:0.159427333247764,0.045);
\draw[draw=black,fill=purple,fill opacity=0.75,line width=0.259723636363636pt] (axis cs:0.159427333247764,0) rectangle (axis cs:0.179031510124395,0.0325);
\draw[draw=black,fill=purple,fill opacity=0.75,line width=0.259723636363636pt] (axis cs:0.179031510124395,0) rectangle (axis cs:0.198635687001027,0.0325);
\draw[draw=black,fill=purple,fill opacity=0.75,line width=0.259723636363636pt] (axis cs:0.198635687001027,0) rectangle (axis cs:0.218239863877658,0.0225);
\draw[draw=black,fill=purple,fill opacity=0.75,line width=0.259723636363636pt] (axis cs:0.218239863877658,0) rectangle (axis cs:0.237844040754289,0.0275);
\draw[draw=black,fill=purple,fill opacity=0.75,line width=0.259723636363636pt] (axis cs:0.237844040754289,0) rectangle (axis cs:0.257448217630921,0.05);
\draw[draw=black,fill=purple,fill opacity=0.75,line width=0.259723636363636pt] (axis cs:0.25744821763092,0) rectangle (axis cs:0.277052394507552,0.0225);
\draw[draw=black,fill=purple,fill opacity=0.75,line width=0.259723636363636pt] (axis cs:0.277052394507552,0) rectangle (axis cs:0.296656571384183,0.0325);
\draw[draw=black,fill=purple,fill opacity=0.75,line width=0.259723636363636pt] (axis cs:0.296656571384183,0) rectangle (axis cs:0.316260748260814,0.0275);
\draw[draw=black,fill=purple,fill opacity=0.75,line width=0.259723636363636pt] (axis cs:0.316260748260814,0) rectangle (axis cs:0.335864925137445,0.0175);
\draw[draw=black,fill=purple,fill opacity=0.75,line width=0.259723636363636pt] (axis cs:0.335864925137445,0) rectangle (axis cs:0.355469102014077,0.0175);
\draw[draw=black,fill=purple,fill opacity=0.75,line width=0.259723636363636pt] (axis cs:0.355469102014077,0) rectangle (axis cs:0.375073278890708,0.0225);
\draw[draw=black,fill=purple,fill opacity=0.75,line width=0.259723636363636pt] (axis cs:0.375073278890708,0) rectangle (axis cs:0.394677455767339,0.0075);
\draw[draw=black,fill=purple,fill opacity=0.75,line width=0.259723636363636pt] (axis cs:0.394677455767339,0) rectangle (axis cs:0.41428163264397,0.0175);
\draw[draw=black,fill=purple,fill opacity=0.75,line width=0.259723636363636pt] (axis cs:0.41428163264397,0) rectangle (axis cs:0.433885809520602,0.01);
\draw[draw=black,fill=purple,fill opacity=0.75,line width=0.259723636363636pt] (axis cs:0.433885809520602,0) rectangle (axis cs:0.453489986397233,0.015);
\draw[draw=black,fill=purple,fill opacity=0.75,line width=0.259723636363636pt] (axis cs:0.453489986397233,0) rectangle (axis cs:0.473094163273864,0.015);
\draw[draw=black,fill=purple,fill opacity=0.75,line width=0.259723636363636pt] (axis cs:0.473094163273864,0) rectangle (axis cs:0.492698340150495,0.0125);
\draw[draw=black,fill=purple,fill opacity=0.75,line width=0.259723636363636pt] (axis cs:0.492698340150495,0) rectangle (axis cs:0.512302517027127,0.0075);
\draw[draw=black,fill=purple,fill opacity=0.75,line width=0.259723636363636pt] (axis cs:0.512302517027127,0) rectangle (axis cs:0.531906693903758,0.0075);
\draw[draw=black,fill=purple,fill opacity=0.75,line width=0.259723636363636pt] (axis cs:0.531906693903758,0) rectangle (axis cs:0.551510870780389,0.0125);
\draw[draw=black,fill=purple,fill opacity=0.75,line width=0.259723636363636pt] (axis cs:0.551510870780389,0) rectangle (axis cs:0.57111504765702,0.015);
\draw[draw=black,fill=purple,fill opacity=0.75,line width=0.259723636363636pt] (axis cs:0.57111504765702,0) rectangle (axis cs:0.590719224533652,0.0025);
\draw[draw=black,fill=purple,fill opacity=0.75,line width=0.259723636363636pt] (axis cs:0.590719224533652,0) rectangle (axis cs:0.610323401410283,0.01);
\draw[draw=black,fill=purple,fill opacity=0.75,line width=0.259723636363636pt] (axis cs:0.610323401410283,0) rectangle (axis cs:0.629927578286914,0.0025);
\draw[draw=black,fill=purple,fill opacity=0.75,line width=0.259723636363636pt] (axis cs:0.629927578286914,0) rectangle (axis cs:0.649531755163545,0.0025);
\draw[draw=black,fill=purple,fill opacity=0.75,line width=0.259723636363636pt] (axis cs:0.649531755163545,0) rectangle (axis cs:0.669135932040177,0.005);
\draw[draw=black,fill=purple,fill opacity=0.75,line width=0.259723636363636pt] (axis cs:0.669135932040177,0) rectangle (axis cs:0.688740108916808,0.005);
\draw[draw=black,fill=purple,fill opacity=0.75,line width=0.259723636363636pt] (axis cs:0.688740108916808,0) rectangle (axis cs:0.708344285793439,0);
\draw[draw=black,fill=purple,fill opacity=0.75,line width=0.259723636363636pt] (axis cs:0.708344285793439,0) rectangle (axis cs:0.72794846267007,0.0175);
\draw[draw=black,fill=purple,fill opacity=0.75,line width=0.259723636363636pt] (axis cs:0.72794846267007,0) rectangle (axis cs:0.747552639546701,0.01);
\draw[draw=black,fill=purple,fill opacity=0.75,line width=0.259723636363636pt] (axis cs:0.747552639546702,0) rectangle (axis cs:0.767156816423333,0.0075);
\draw[draw=black,fill=purple,fill opacity=0.75,line width=0.259723636363636pt] (axis cs:0.767156816423333,0) rectangle (axis cs:0.786760993299964,0.0275);
\draw[draw=black,fill=purple,fill opacity=0.75,line width=0.259723636363636pt] (axis cs:0.786760993299964,0) rectangle (axis cs:0.806365170176595,0.0125);
\draw[draw=black,fill=purple,fill opacity=0.75,line width=0.259723636363636pt] (axis cs:0.806365170176595,0) rectangle (axis cs:0.825969347053226,0.0075);
\draw[draw=black,fill=purple,fill opacity=0.75,line width=0.259723636363636pt] (axis cs:0.825969347053227,0) rectangle (axis cs:0.845573523929858,0.0125);
\draw[draw=black,fill=purple,fill opacity=0.75,line width=0.259723636363636pt] (axis cs:0.845573523929858,0) rectangle (axis cs:0.865177700806489,0.0175);
\draw[draw=black,fill=purple,fill opacity=0.75,line width=0.259723636363636pt] (axis cs:0.865177700806489,0) rectangle (axis cs:0.88478187768312,0.0075);
\draw[draw=black,fill=purple,fill opacity=0.75,line width=0.259723636363636pt] (axis cs:0.88478187768312,0) rectangle (axis cs:0.904386054559751,0.0275);
\draw[draw=black,fill=purple,fill opacity=0.75,line width=0.259723636363636pt] (axis cs:0.904386054559752,0) rectangle (axis cs:0.923990231436383,0.005);
\draw[draw=black,fill=purple,fill opacity=0.75,line width=0.259723636363636pt] (axis cs:0.923990231436383,0) rectangle (axis cs:0.943594408313014,0.03);
\draw[draw=black,fill=purple,fill opacity=0.75,line width=0.259723636363636pt] (axis cs:0.943594408313014,0) rectangle (axis cs:0.963198585189645,0.01);
\draw[draw=black,fill=purple,fill opacity=0.75,line width=0.259723636363636pt] (axis cs:0.963198585189645,0) rectangle (axis cs:0.982802762066277,0.0075);
\end{axis}

\end{tikzpicture}}
    \resizebox{0.4\linewidth}{0.35\linewidth}{
\begin{tikzpicture}

\definecolor{darkgray176}{RGB}{176,176,176}
\definecolor{steelblue31119180}{RGB}{31,119,180}

\begin{axis}[
tick align=outside,
tick pos=left,
title={Nonconformity Scores for Decision (Table-top)},
x grid style={darkgray176},
xlabel={Softmax},
xmin=-0.02, xmax=1.02,
xtick style={color=black},
y grid style={darkgray176},
label style={font=\Large},
ymin=0, ymax=0.1,
ytick style={color=black},
label style={font=\Large},
yticklabel style={
  /pgf/number format/precision=3,
  /pgf/number format/fixed}
]
\draw[draw=black,fill=steelblue31119180,fill opacity=0.75,line width=0.259723636363634pt] (axis cs:4.50036384454407e-05,0) rectangle (axis cs:0.0200441035656765,0.1378125);
\draw[draw=black,fill=steelblue31119180,fill opacity=0.75,line width=0.259723636363634pt] (axis cs:0.0200441035656765,0) rectangle (axis cs:0.0400432034929076,0.0465625);
\draw[draw=black,fill=steelblue31119180,fill opacity=0.75,line width=0.259723636363634pt] (axis cs:0.0400432034929076,0) rectangle (axis cs:0.0600423034201387,0.0325);
\draw[draw=black,fill=steelblue31119180,fill opacity=0.75,line width=0.259723636363634pt] (axis cs:0.0600423034201387,0) rectangle (axis cs:0.0800414033473698,0.03375);
\draw[draw=black,fill=steelblue31119180,fill opacity=0.75,line width=0.259723636363634pt] (axis cs:0.0800414033473698,0) rectangle (axis cs:0.100040503274601,0.03);
\draw[draw=black,fill=steelblue31119180,fill opacity=0.75,line width=0.259723636363634pt] (axis cs:0.100040503274601,0) rectangle (axis cs:0.120039603201832,0.0296875);
\draw[draw=black,fill=steelblue31119180,fill opacity=0.75,line width=0.259723636363634pt] (axis cs:0.120039603201832,0) rectangle (axis cs:0.140038703129063,0.0240625);
\draw[draw=black,fill=steelblue31119180,fill opacity=0.75,line width=0.259723636363634pt] (axis cs:0.140038703129063,0) rectangle (axis cs:0.160037803056294,0.0265625);
\draw[draw=black,fill=steelblue31119180,fill opacity=0.75,line width=0.259723636363634pt] (axis cs:0.160037803056294,0) rectangle (axis cs:0.180036902983525,0.02875);
\draw[draw=black,fill=steelblue31119180,fill opacity=0.75,line width=0.259723636363634pt] (axis cs:0.180036902983525,0) rectangle (axis cs:0.200036002910756,0.029375);
\draw[draw=black,fill=steelblue31119180,fill opacity=0.75,line width=0.259723636363634pt] (axis cs:0.200036002910756,0) rectangle (axis cs:0.220035102837987,0.0315625);
\draw[draw=black,fill=steelblue31119180,fill opacity=0.75,line width=0.259723636363634pt] (axis cs:0.220035102837987,0) rectangle (axis cs:0.240034202765219,0.025625);
\draw[draw=black,fill=steelblue31119180,fill opacity=0.75,line width=0.259723636363634pt] (axis cs:0.240034202765219,0) rectangle (axis cs:0.26003330269245,0.025);
\draw[draw=black,fill=steelblue31119180,fill opacity=0.75,line width=0.259723636363634pt] (axis cs:0.26003330269245,0) rectangle (axis cs:0.280032402619681,0.034375);
\draw[draw=black,fill=steelblue31119180,fill opacity=0.75,line width=0.259723636363634pt] (axis cs:0.280032402619681,0) rectangle (axis cs:0.300031502546912,0.0278125);
\draw[draw=black,fill=steelblue31119180,fill opacity=0.75,line width=0.259723636363634pt] (axis cs:0.300031502546912,0) rectangle (axis cs:0.320030602474143,0.0075);
\draw[draw=black,fill=steelblue31119180,fill opacity=0.75,line width=0.259723636363634pt] (axis cs:0.320030602474143,0) rectangle (axis cs:0.340029702401374,0.00625);
\draw[draw=black,fill=steelblue31119180,fill opacity=0.75,line width=0.259723636363634pt] (axis cs:0.340029702401374,0) rectangle (axis cs:0.360028802328605,0.0071875);
\draw[draw=black,fill=steelblue31119180,fill opacity=0.75,line width=0.259723636363634pt] (axis cs:0.360028802328605,0) rectangle (axis cs:0.380027902255836,0.006875);
\draw[draw=black,fill=steelblue31119180,fill opacity=0.75,line width=0.259723636363634pt] (axis cs:0.380027902255836,0) rectangle (axis cs:0.400027002183067,0.004375);
\draw[draw=black,fill=steelblue31119180,fill opacity=0.75,line width=0.259723636363634pt] (axis cs:0.400027002183067,0) rectangle (axis cs:0.420026102110298,0.005625);
\draw[draw=black,fill=steelblue31119180,fill opacity=0.75,line width=0.259723636363634pt] (axis cs:0.420026102110298,0) rectangle (axis cs:0.44002520203753,0.0075);
\draw[draw=black,fill=steelblue31119180,fill opacity=0.75,line width=0.259723636363634pt] (axis cs:0.440025202037529,0) rectangle (axis cs:0.460024301964761,0.0078125);
\draw[draw=black,fill=steelblue31119180,fill opacity=0.75,line width=0.259723636363634pt] (axis cs:0.460024301964761,0) rectangle (axis cs:0.480023401891992,0.00875);
\draw[draw=black,fill=steelblue31119180,fill opacity=0.75,line width=0.259723636363634pt] (axis cs:0.480023401891992,0) rectangle (axis cs:0.500022501819223,0.0115625);
\draw[draw=black,fill=steelblue31119180,fill opacity=0.75,line width=0.259723636363634pt] (axis cs:0.500022501819223,0) rectangle (axis cs:0.520021601746454,0.01375);
\draw[draw=black,fill=steelblue31119180,fill opacity=0.75,line width=0.259723636363634pt] (axis cs:0.520021601746454,0) rectangle (axis cs:0.540020701673685,0.009375);
\draw[draw=black,fill=steelblue31119180,fill opacity=0.75,line width=0.259723636363634pt] (axis cs:0.540020701673685,0) rectangle (axis cs:0.560019801600916,0.009375);
\draw[draw=black,fill=steelblue31119180,fill opacity=0.75,line width=0.259723636363634pt] (axis cs:0.560019801600916,0) rectangle (axis cs:0.580018901528147,0.014375);
\draw[draw=black,fill=steelblue31119180,fill opacity=0.75,line width=0.259723636363634pt] (axis cs:0.580018901528147,0) rectangle (axis cs:0.600018001455378,0.009375);
\draw[draw=black,fill=steelblue31119180,fill opacity=0.75,line width=0.259723636363634pt] (axis cs:0.600018001455378,0) rectangle (axis cs:0.620017101382609,0.015);
\draw[draw=black,fill=steelblue31119180,fill opacity=0.75,line width=0.259723636363634pt] (axis cs:0.620017101382609,0) rectangle (axis cs:0.64001620130984,0.0153125);
\draw[draw=black,fill=steelblue31119180,fill opacity=0.75,line width=0.259723636363634pt] (axis cs:0.64001620130984,0) rectangle (axis cs:0.660015301237071,0.015625);
\draw[draw=black,fill=steelblue31119180,fill opacity=0.75,line width=0.259723636363634pt] (axis cs:0.660015301237072,0) rectangle (axis cs:0.680014401164303,0.01125);
\draw[draw=black,fill=steelblue31119180,fill opacity=0.75,line width=0.259723636363634pt] (axis cs:0.680014401164303,0) rectangle (axis cs:0.700013501091534,0.0109375);
\draw[draw=black,fill=steelblue31119180,fill opacity=0.75,line width=0.259723636363634pt] (axis cs:0.700013501091534,0) rectangle (axis cs:0.720012601018765,0.011875);
\draw[draw=black,fill=steelblue31119180,fill opacity=0.75,line width=0.259723636363634pt] (axis cs:0.720012601018765,0) rectangle (axis cs:0.740011700945996,0.0128125);
\draw[draw=black,fill=steelblue31119180,fill opacity=0.75,line width=0.259723636363634pt] (axis cs:0.740011700945996,0) rectangle (axis cs:0.760010800873227,0.00625);
\draw[draw=black,fill=steelblue31119180,fill opacity=0.75,line width=0.259723636363634pt] (axis cs:0.760010800873227,0) rectangle (axis cs:0.780009900800458,0.005625);
\draw[draw=black,fill=steelblue31119180,fill opacity=0.75,line width=0.259723636363634pt] (axis cs:0.780009900800458,0) rectangle (axis cs:0.800009000727689,0.003125);
\draw[draw=black,fill=steelblue31119180,fill opacity=0.75,line width=0.259723636363634pt] (axis cs:0.800009000727689,0) rectangle (axis cs:0.82000810065492,0.0046875);
\draw[draw=black,fill=steelblue31119180,fill opacity=0.75,line width=0.259723636363634pt] (axis cs:0.82000810065492,0) rectangle (axis cs:0.840007200582151,0.0046875);
\draw[draw=black,fill=steelblue31119180,fill opacity=0.75,line width=0.259723636363634pt] (axis cs:0.840007200582151,0) rectangle (axis cs:0.860006300509382,0.0046875);
\draw[draw=black,fill=steelblue31119180,fill opacity=0.75,line width=0.259723636363634pt] (axis cs:0.860006300509382,0) rectangle (axis cs:0.880005400436613,0.003125);
\draw[draw=black,fill=steelblue31119180,fill opacity=0.75,line width=0.259723636363634pt] (axis cs:0.880005400436613,0) rectangle (axis cs:0.900004500363844,0.0034375);
\draw[draw=black,fill=steelblue31119180,fill opacity=0.75,line width=0.259723636363634pt] (axis cs:0.900004500363845,0) rectangle (axis cs:0.920003600291076,0.0075);
\draw[draw=black,fill=steelblue31119180,fill opacity=0.75,line width=0.259723636363634pt] (axis cs:0.920003600291076,0) rectangle (axis cs:0.940002700218307,0.0084375);
\draw[draw=black,fill=steelblue31119180,fill opacity=0.75,line width=0.259723636363634pt] (axis cs:0.940002700218307,0) rectangle (axis cs:0.960001800145538,0.025);
\draw[draw=black,fill=steelblue31119180,fill opacity=0.75,line width=0.259723636363634pt] (axis cs:0.960001800145538,0) rectangle (axis cs:0.980000900072769,0.02875);
\draw[draw=black,fill=steelblue31119180,fill opacity=0.75,line width=0.259723636363634pt] (axis cs:0.980000900072769,0) rectangle (axis cs:1,0.10875);
\end{axis}

\end{tikzpicture}}
    \caption{The left and right figures depict nonconformity distributions for perception and plan generation in table-top manipulation.}
    \label{fig: arm-unc-dist}
\end{figure}
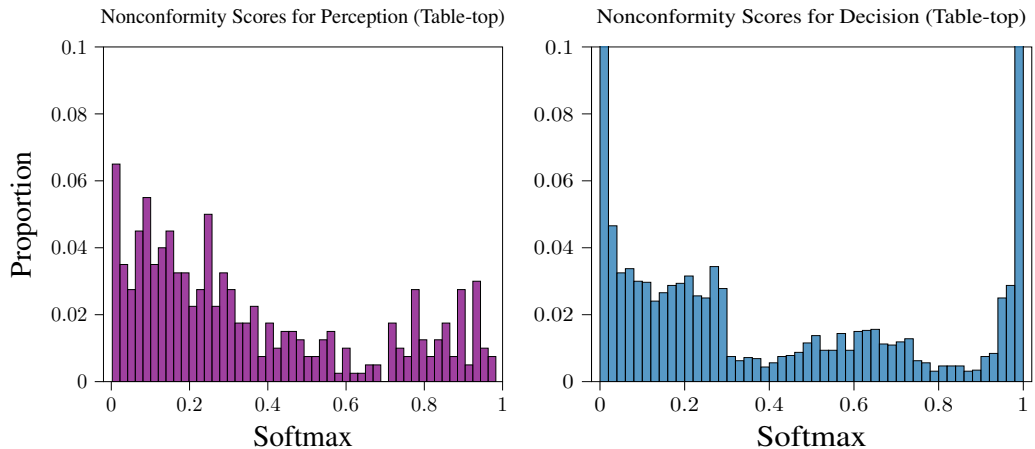

\begin{figure}[t]
    \centering
    \includegraphics[width=\linewidth]{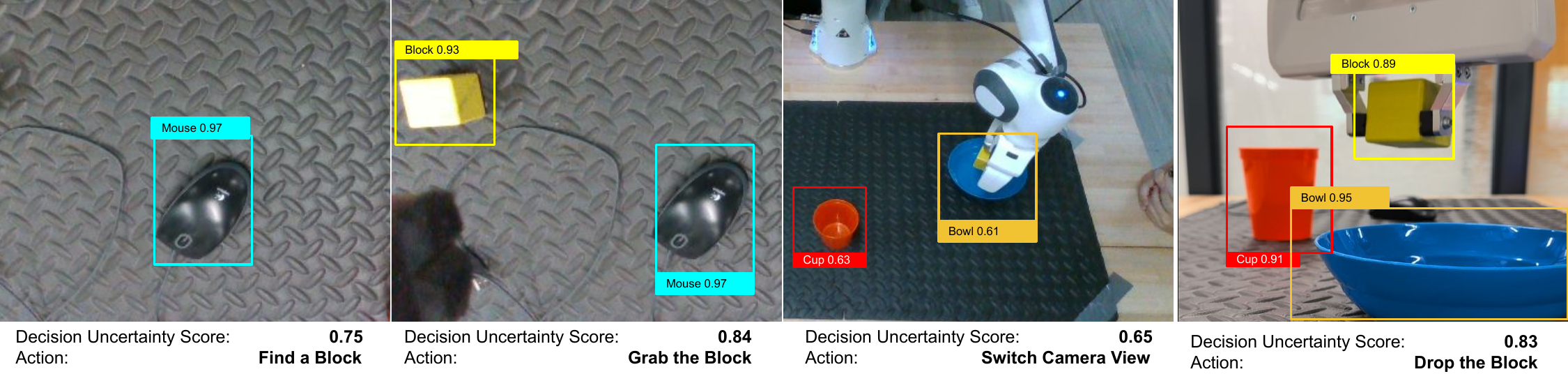}
    \caption{Illustration of our strategy in Fig. \ref{fig: strategy} solving table-top manipulation tasks.}
    \label{fig: use-cases-arm}
\end{figure}

Beyond autonomous driving, we demonstrate our planning strategy with uncertainty disentanglement through a set of table-top manipulation tasks. 

For table-top manipulation, we obtain nonconformity distributions (presented in Fig. \ref{fig: unc-dist}).  We collect $622$ images from three camera views for calibration and supply the images to LLaVA with task descriptions such as ``[pick/place] [an object] on the [table]"). We define a set of specifications and present the nonconformity distributions in Fig. \ref{fig: arm-unc-dist}. 

Fig. \ref{fig: use-cases-arm} shows a use case where we successfully complete a task ``drop the block into a bowl'' while satisfying all the defined specifications. The specifications include

$\phi_1 = \lalways (\neg \text{grab the cup} \land \neg \text{grab the bowl})$,

$\phi_2 = \lalways ( \neg \text{block} \rightarrow (\neg \text{grab the block} \land \text{locate the block} ) )$,

$\phi_3 = \lalways ( \neg \text{mouse} \rightarrow \neg \text{grab the mouse} \land \text{locate the mouse} )$,

$\phi_4 = \lalways ( \neg \text{cup} \rightarrow \text{locate the cup} )$,

$\phi_5 = \lalways ( \neg \text{bowl} \rightarrow \text{locate the bowl} )$.

This use case indicates the generalizability of our planning strategy across multiple domains.

\subsection{Effect of User-specified Thresholds for Uncertainty Levels}
\label{app: thresholds}

The thresholds---$t_p$ and $t_d$---are user-defined theoretical lower bounds on perception and decision errors, indicating the user's error tolerance. We note that different thresholds may lead to different performance. To investigate this further, we use our fine-tuned model to present the effect of different thresholds when deployed in the Carla simulations.

\begin{table}[t]
    \centering
    \begin{tabular}{lccccc}
    \toprule
        Threshold ($t_d = t_p$) & 0.5 & 0.6 & 0.7 & 0.8 & 0.9 \\
    \midrule
        Accuracy & 0.632 & 0.789 & 0.842 & 0.891 & 0.947 \\
        AS Frequency & 0.294 & 0.529 & 1.176 & 2.117 & 3.647 \\
        Satisfy Prob & 0.857 & 0.904 & 0.959 & 0.967 & 0.981 \\
    \bottomrule
    \end{tabular}
    \caption{Effect of different thresholds on the fine-tuned model performance.}
    \label{tab: threshold_study}
\end{table}

Note that in Tab. \ref{tab: threshold_study}, threshold refers to the value for both $t_p$ and $t_d$, e.g., $t_p = t_d = 0.5$. Accuracy refers to the average probability of correctly classifying objects, and AS frequency refers to the average number of active sensing interventions triggered to meet the threshold, i.e., the number of iterations. Lastly, Satisfy Prob refers to the average probability of satisfying the specifications.

There is a trade-off between the perception/decision accuracy and the frequency of triggering active sensing (i.e., computation overhead).


\end{document}